\documentclass[10pt,journal,compsoc]{IEEEtran}

\usepackage{times}
\usepackage{epsfig}
\usepackage{graphicx}
\usepackage{amsmath}
\usepackage{amssymb}
\usepackage{booktabs} 
\usepackage{array} 
\usepackage{paralist} 
\usepackage{verbatim} 

\usepackage{stmaryrd,bm,abraces}
\usepackage{color}
\usepackage{float}
\usepackage{array}
\usepackage{makecell}
\usepackage{multirow}
\usepackage{verbatim}
\usepackage{soul}
\usepackage{float}
\usepackage{enumitem}
\usepackage{arydshln}
\usepackage{mathtools}
\usepackage{threeparttable}
\usepackage[lined,ruled,commentsnumbered]{algorithm2e}
\usepackage{lipsum}
\usepackage{afterpage}
\usepackage{xcolor}
\usepackage{caption}
\usepackage{subcaption}
\usepackage{xcolor}
\usepackage{tabu}

\usepackage{environ}
\NewEnviron{lequation}{%
\begin{equation}
\scalebox{1.1}{$\BODY$}
\end{equation}
}
\newcommand{\nbits}{b}
\newcommand{\vp}{\varoplus}
\newcommand{\vm}{\varominus}
\newcommand{\xhat}{{\hat{{x}}}}
\newcommand{\bigPhi}{ \boldsymbol{\hat{\Phi}} }
\newcommand{\x}{{{x}}}
\newcommand{\C}{\mathcal{C}}
\newcommand{\D}{\mathcal{D}}

\newcommand{\mihash}{$\mathsf{MIHash}$}
\newcommand{\eg}{\textit{e.g.}\ }
\newcommand{\ie}{\textit{i.e.}\ }
\newcommand{\etal}{\textit{et al.}\ }
\newcommand{\nus}[1]{\textit{nus-#1}}
\newcommand{\cifar}[1]{\textit{cifar-#1}}

\newcommand{\vggf}{{VGG-F}}
\newcommand{\gist}{{GIST}}
\newcommand{\alexnet}{{AlexNet}}
\newcommand{\f}[1]{\textcolor{black}{#1}}

\ifCLASSOPTIONcompsoc
  \usepackage[nocompress]{cite}
\else
  \usepackage{cite}
\fi
\ifCLASSINFOpdf
\else
\fi

\newcolumntype{v}{>{\centering\arraybackslash}m{1.07cm}}
\newcolumntype{y}{>{\centering\arraybackslash}m{0.05cm}}
\newcolumntype{L}{>{\centering\arraybackslash}m{2.3cm}}
\newcolumntype{X}{>{\centering\arraybackslash}m{3.2cm}}
\newcolumntype{Y}{>{\centering\arraybackslash}m{2.7cm}}
\newcolumntype{S}{>{\centering\arraybackslash}m{3.0cm}}
\newcolumntype{F}{>{\centering\arraybackslash}m{4.0cm}}
\newcolumntype{K}{>{\centering\arraybackslash}m{3.7cm}}
\newcolumntype{N}{>{\centering\arraybackslash}m{0.6cm}}
\newcolumntype{Z}{>{\centering\arraybackslash}m{0.1cm}}
\newcolumntype{w}{>{\centering\arraybackslash}m{1.9cm}}
\newcolumntype{p}{>{\centering\arraybackslash}m{1.2cm}}

\begin{document}
%
\title{Hashing with Mutual Information}
%
%
%
%

\author{Fatih~Cakir*,~\IEEEmembership{Member,~IEEE,}
        Kun~He*,~\IEEEmembership{Student Member,~IEEE,}
        Sarah~Adel~Bargal,~\IEEEmembership{Student Member,~IEEE}
        and~Stan~Sclaroff,~\IEEEmembership{Fellow,~IEEE}
\IEEEcompsocitemizethanks{\IEEEcompsocthanksitem F. Cakir was with the Department
of Computer Science, Boston University, Boston, MA, 02215.\protect\\
E-mail: fcakir@bu.edu
\IEEEcompsocthanksitem K. He, S. A. Bargal, and S. Sclaroff are with the Department
of Computer Science, Boston University, Boston, MA, 02215.\protect\\
E-mails: {hekun, sbargal, sclaroff}@bu.edu
\IEEEcompsocthanksitem[*] The first two authors contributed equally.
}
}

%
%

\markboth{IEEE Transactions on Pattern Analysis and Machine Intelligence}%
{Cakir \MakeLowercase{\textit{et al.}}: Hashing with Mutual Information}
%



\IEEEtitleabstractindextext{%
\begin{abstract}
Binary vector embeddings enable fast nearest neighbor retrieval in large databases of high-dimensional objects, and play an important role in many practical applications, such as image and video retrieval.
We study the problem of learning binary vector embeddings under a supervised setting, also known as hashing.
We propose a novel supervised hashing method based on optimizing an information-theoretic quantity, \emph{mutual information}. 
We show that optimizing mutual information can reduce ambiguity in the induced neighborhood structure in the learned Hamming space, which is essential in obtaining high retrieval performance.
To this end, we optimize mutual information in deep neural networks with minibatch stochastic gradient descent,  with a formulation that maximally and efficiently utilizes available supervision.
Experiments on four image retrieval benchmarks, including ImageNet, confirm the effectiveness of our method in learning high-quality binary embeddings for nearest neighbor retrieval.
\end{abstract}

}

\maketitle


\IEEEdisplaynontitleabstractindextext

%
\IEEEpeerreviewmaketitle

\ifCLASSOPTIONcompsoc
\IEEEraisesectionheading{\section{Introduction}\label{sec:introduction}}
\else
\section{Introduction}
\label{sec:introduction}
\fi

%
%
%
%
\IEEEPARstart{I}{n} computer vision and many other application areas, there is typically an abundance of data with high-dimensional raw representations, such as mega-pixel images and high-definition videos.
Besides obvious storage challenges, high-dimensional data pose additional challenges for semantic-level processing and understanding. One prominent example that we focus on in this paper is nearest neighbor retrieval.
In applications such as image and video search, person and object recognition in photo collections, and action detection and classification in surveillance video, it is often necessary to map high-dimensional data objects to low-dimensional vector representations to allow for efficient retrieval of similar instances in large databases.
In addition, the desired semantic similarity can vary from  task to task, often prescribed by available supervision, \eg\ class labels.
Therefore, the mapping process is also responsible for leveraging supervised learning to encode task-specific similarities, such that objects that are semantically similar are mapped to close neighbors in the resulting vector space.

In this paper, we consider the problem of {learning} low-dimensional \emph{binary} vector embeddings of high-dimensional data, also known as hashing.
Binary embeddings enjoy a small memory footprint and permit fast search mechanisms, as Hamming distance computation between binary vectors can be implemented very efficiently in modern CPUs. 
As a result, across a variety of domains, hashing approaches have  been widely utilized in applications requiring fast (approximate) nearest neighbor retrieval.
Examples include: image annotation \cite{imret}, visual tracking \cite{visualtrack}, 3D reconstruction \cite{Cheng2014CVPR}, video segmentation \cite{XiaoVideoSegCVPR14}, object detection \cite{Dean100KCVPR13}, audio search \cite{Wang03shazam}, multimedia retrieval \cite{hashing-video-retrieval,Song-scalable-mul-retrieval,Gao-scalable-mul-retrieval}, and large-scale clustering \cite{Haveliwala00scalabletechniques,FergusHashClustering2015, LSHClustering}.
Our goal is to learn hashing functions that can result in optimal nearest neighbor retrieval performance.
In particular, as motivated above, we approach hashing as a {supervised} learning problem, such that the learned binary embeddings encode task-specific semantic similarity.

Supervised hashing is a well-studied problem. Although many different formulations exist,  all supervised hashing formulations essentially constrain the learned Hamming distance to agree with the given supervision.
Such supervision can be specified as pairwise affinity labels: pairs of objects are annotated with binary labels indicating their pairwise relationships as either ``similar'' or ``dissimilar.''
In this case, a common learning strategy is \emph{affinity matching}: the learned binary embedding should evaluate to low Hamming distances between similar pairs, and high Hamming distances between dissimilar pairs.
Alternatively, supervision can also be given in terms of local relative distance comparisons, most notably three-tuples of examples, or ``triplets", where one example is constrained to have a smaller distance to the second example than the third. This can be termed \emph{local ranking}.
Typically,  for ease of optimization, 
these formulations define and optimize loss functions that match the form of supervision, \emph{i.e.}, defined on training pairs or triplets. 
However, loss functions used in affinity matching and local ranking methods are usually only indirectly related to retrieval performance, and in order to optimize overall retrieval performance, it is often necessary to introduce additional regularization terms, or parameters such as margins, thresholds, and scaling factors.

We argue that approaches such as affinity matching and local ranking are insufficient to achieve optimal nearest neighbor retrieval performance.
Instead, we view supervised hashing through an information-theoretic lens, and propose a novel solution tailored for the task of nearest neighbor retrieval.
Our key observation is that a good binary embedding should well separate neighbors and non-neighbors in the Hamming space, or, achieve low \emph{neighborhood ambiguity}.
An alternative viewpoint is that the learned Hamming embedding should carry a high amount of information regarding the desired neighborhood structure.
To quantify neighborhood ambiguity, we use a well-known quantity from information theory, \emph{mutual information}, and 
show that it has direct and strong correlations with standard ranking-based retrieval performance metrics.
An appealing property of the mutual information objective is that it is free of tuning parameters, unlike others that may require thresholds, margins, and so on.
Finally, to optimize mutual information, we  relax the original NP-hard discrete optimization problem, 
and develop a gradient-based optimization framework that can be efficiently applied with minibatch stochastic gradient descent in deep neural networks.

To briefly summarize our contributions, we propose a novel supervised hashing method that is based on quantifying and minimizing neighborhood ambiguity in the learned Hamming space, using mutual information as the learning objective. An end-to-end gradient-based optimization framework is developed, with an efficient minibatch-based implementation.
Our proposed hashing method is named \mihash.\footnote{Our MATLAB implementation of \mihash\ is publicly available at \texttt{https://github.com/fcakir/mihash}}
We conduct image retrieval experiments on four standard  benchmarks: CIFAR-10, NUSWIDE, LabelMe, and ImageNet.
\mihash\ achieves state-of-the-art retrieval performance across all datasets.

This paper builds upon the formulations introduced in \cite{Cakir_2017_ICCV} which primarily addressed \emph{online hashing}, a separate and distinct task within the “hashing for approximate nearest neighbor retrieval” problem domain. 
In this paper, we expand and improve the formulation introduced in \cite{Cakir_2017_ICCV}, and propose a hashing method for the \emph{batch learning} setting. 
Notably, our proposed formulation is amenable for deep learning, and we propose an efficient formulation for utilizing supervision in minibatches during stochastic optimization. 
We further conduct extensive experiments in the batch learning setup, and provide detailed empirical analysis for our proposed approach.

The rest of this paper is organized as follows. First, the relevant literature is reviewed in Section~\ref{sec:relatedwork}. 
We propose and analyze mutual information as a learning objective for hashing in Section~\ref{sec:formulation}, and then discuss its optimization using stochastic gradient descent and deep neural networks in Section~\ref{sec:optim}.
Section~\ref{sec:exp} presents experimental results and empirical analysis of the proposed algorithm's behavior. Finally, Section~\ref{sec:conclusion} presents the conclusions.

\section{Related Work}
\label{sec:relatedwork}

Many hashing methods have been introduced over the years. 
While providing a precise taxonomy of the literature is difficult, a rough grouping can be made as data-independent and dependent techniques. 
Data-independent techniques do not exploit the data distribution during hashing. Instead, similarity, as induced by a particular metric, is often preserved. This is achieved by maximizing the probability of ``collision'' when hashing similar items. Notable earlier examples include Locality Sensitive Hashing methods \cite{LSH, LSH1, KLSH} where distance functions such as the Euclidean, Jaccard, and Cosine distances are approximated. These methods usually have theoretical guarantees on the approximation  quality and conform with sub-linear retrieval mechanisms. However, they are confined to certain metrics, and they ignore the data distribution and accompanying supervision. 

In contrast to data-independent techniques, recent approaches are data-dependent such that hash mappings are learned from the training set. 
While empirical evidence for the superiority of these methods over their data-independent counterparts is plentiful in the literature, a recent study has also theoretically validated their performance advantage \cite{Andoni2015}.
These methods can be considered as binary embeddings that map the data into Hamming space while preserving a specific neighborhood structure. 
Such a neighborhood structure can be  derived from meta-data (\textit{e.g.}, labels), or can be completely determined by the user (\textit{e.g.}, via similarity-dissimilarity indicators of data pairs). 
With the binary embeddings, distances can be very efficiently computed, thereby allowing even a linear search to be done efficiently for large-scale corpora. 
These data-dependent methods can be grouped as follows: similarity preserving techniques \cite{SSH, BRE, SHK, FastHash, SH, LDAHASH, STH, Ge2014,Shen_CVPR2015}, quantization methods \cite{PQ, ITQ, KmeansHashing} and recently, deep learning based methods \cite{Xia_AAAI2014,Lin_CVPRW2015, Liong_CVPR2015, Carreira-Perpinan_2015_CVPR,Lai_CVPR2015, Zhao_CVPR2015,Zhuang_CVPR2016, Liu_2016_CVPR, Ziming_VeryDeep_CVPR2016}. 
We now review a few of the prominent techniques in each category. 
For a more comprehensive survey of the hashing literature, we refer readers to \cite{hash_survey_pami_2018}.

\textbf{Quantization methods} are the first group of data-dependent hashing methods.
Such techniques do not assume the availability of supervision, and generally optimize objectives involving a reconstruction error. 
Among these, Semi-Supervised Hashing \cite{SSH} learns the hash mapping by maximizing the empirical accuracy on labeled data and also the entropy of the generated hash functions on unlabeled data.  
This is shown to be very similar to doing a PCA analysis where the hash functions are the eigenvectors of a covariance matrix. 
Other noteworthy work includes PCA-inspired methods where the principal components are taken as the hash functions. If ``groups'' that are suitable for clustering exist within the data,  then further refining the principal components for better binarization has shown to be beneficial, as in Iterative Quantization \cite{ITQ} and K-means Hashing \cite{KmeansHashing}. 

Unsupervised quantization can also be approached as a special case of generative modeling.
Semantic Hashing \cite{SemanticHashing} is one early example that is based on the autoencoding principle.
It learns a generative model to encode data, in the form of stacked Restricted Boltzmann Machines (RBMs).
Carreira-Perpinan and Raziperchikolaei \cite{Carreira-Perpinan_2015_CVPR} propose Binary Autoencoders, and construct autoencoders with a binary latent layer. 
They argue that finding the hash mapping without relaxing the binary constraints will yield  better solutions, while in relaxation approaches that are more common in the literature, quantization errors can degrade the quality of learned hash functions. 
More recently, Stochastic Generative Hashing \cite{SGH} learns a generative hashing model based on the minimum description length principle, and uses stochastic distributional gradient descent to optimize the associated variational objective and handle the difficulty in having binary stochastic neurons.

\textbf{Similarity preserving methods}, on the other hand, aim to construct binary embeddings that optimize loss functions induced from the supervision provided. Both the affinity matching and local ranking methods mentioned in Section~\ref{sec:introduction} belong to this group. Among such techniques, Minimal Loss Hashing \cite{MLH} considers minimizing a hinge-like loss function motivated by Structural SVMs \cite{SVMstruct}. In Binary Reconstructive Embeddings \cite{BRE}, a kernel-based solution is proposed where the goal is to construct hash functions by minimizing an empirical loss between the input and Hamming space distances via a coordinate descent algorithm. 
Supervised Hashing with Kernels \cite{SHK} also proposes a kernel-based solution; but, instead of preserving the equivalence of the input and Hamming space distances, the kernel function weights are learned by minimizing an objective function based on binary code inner products. 
Spectral Hashing \cite{SH} and Self-Taught Hashing \cite{STH} are other notable lines of work where the similarity of the instances is preserved by solving a graph Laplacian problem. 
Rank alignment methods \cite{SSC, Ding_2015_ICCV} that learn a hash mapping to preserve rankings in the data can also be considered in this group. 

Lately, several ``two-stage'' similarity preserving techniques have also been  proposed, where the learning is decomposed into two steps: binary inference and hash function learning. The binary inference step yields hash codes that best preserve the similarity. These hash codes are used as target vectors in the subsequent hash function learning step, for example, by learning binary classifiers to produce each bit. Notable two-stage methods include Fast Hashing with Decision Trees \cite{TSH, FastHash}, Structured Learning of Binary Codes \cite{LinIJCV2016} and Supervised Discrete Hashing \cite{Shen_CVPR2015}.
All of these similarity preserving methods assume some type of supervision, such as labels or similarity indicators. Thus, in the literature, such techniques are also regarded as supervised hashing solutions. 

\textbf{Deep hashing methods} have recently gained significant prominence following the success of deep neural networks in related tasks such as image classification.
Although hashing methods that employ deep learning can be based on either unsupervised quantization or supervised learning, most existing deep hashing methods are supervised. 
A deep hashing study typically involves a novel architecture, a loss function or a binary inference formulation. 
Among such methods, Lai \etal \cite{Lai_CVPR2015} jointly learn the hash mapping and image features with a triplet loss formulation. This triplet loss ensures that an image is more similar to the second image than to a third one with respect to their binary codes. In \cite{NIN}, a network-in-network (NIN) deep net architecture is used, with a divide-conquer module that is shown to reduce redundancy in the hash bits. 
In \cite{Ziming_VeryDeep_CVPR2016}, the authors follow the work of \cite{Shen_CVPR2015} and \cite{Carreira-Perpinan_2015_CVPR}. Similar to \cite{Shen_CVPR2015} they propose learning the hash mapping by optimizing a classification objective. 
Differently, they consider using a deep net consisting only of fully-connected layers, and use auxiliary variables, as in \cite{Carreira-Perpinan_2015_CVPR}, to circumvent the vanishing gradient problem. 
Deep learning based hashing studies have also proposed sampling pairs or triplets of data instances to learn the hash mapping. Notable examples include \cite{Li_IJCAI2016} and \cite{Wang_ACCV2016}, which optimize a likelihood function which ensures that similar (non-similar) pairs or triplets are mapped to nearby (distant) binary embeddings. 

As the ultimate goal of hashing is to preserve a neighborhood structure in the Hamming space, we propose an information-theoretic solution and directly quantify the neighborhood ambiguity of the generated binary embeddings using a mutual information based criterion. 
Information-theoretic measures have also been considered in past hashing studies. 
Notably, in \cite{Spec_hash_cvpr_2010} an affinity matching formulation is proposed with a pairwise cross-entropy loss to penalize the discrepancy between pairwise Hamming similarities and the ground truth affinities. 
A similar cross-entropy loss is adopted by Zhu \etal \cite{DHN_aaai_2016} in convolutional neural networks.
To permit gradient based optimization, the binary embeddings are relaxed to continuous values and a quantization loss is added. 
Venkatesware \etal \cite{DHN_da_cvpr_2017} also consider the cross-entropy loss in an unsupervised domain adaptation setting.
These simple affinity matching methods are different from our proposed solution, 
where we employ an information-theoretic measure to directly minimize neighborhood ambiguity: separating distance distributions between queries and their neighbor and non-neighbor sets. 
Our proposed mutual information objective is efficient to compute, amenable to batch learning, and leads to state-of-the-art results in standard retrieval benchmarks. 

We utilize a recent study, \cite{ustinova2016learning}, when optimizing our mutual information based objective, and use their differentiable histogram binning technique as a foundation in deriving gradient-based optimization rules.
Note that both our problem setup and objective function are quite different from \cite{ustinova2016learning}. 

\begin{figure*}[t]
\centering
\includegraphics[width=.97\linewidth, trim= 1em 33em 0.3em 1em]{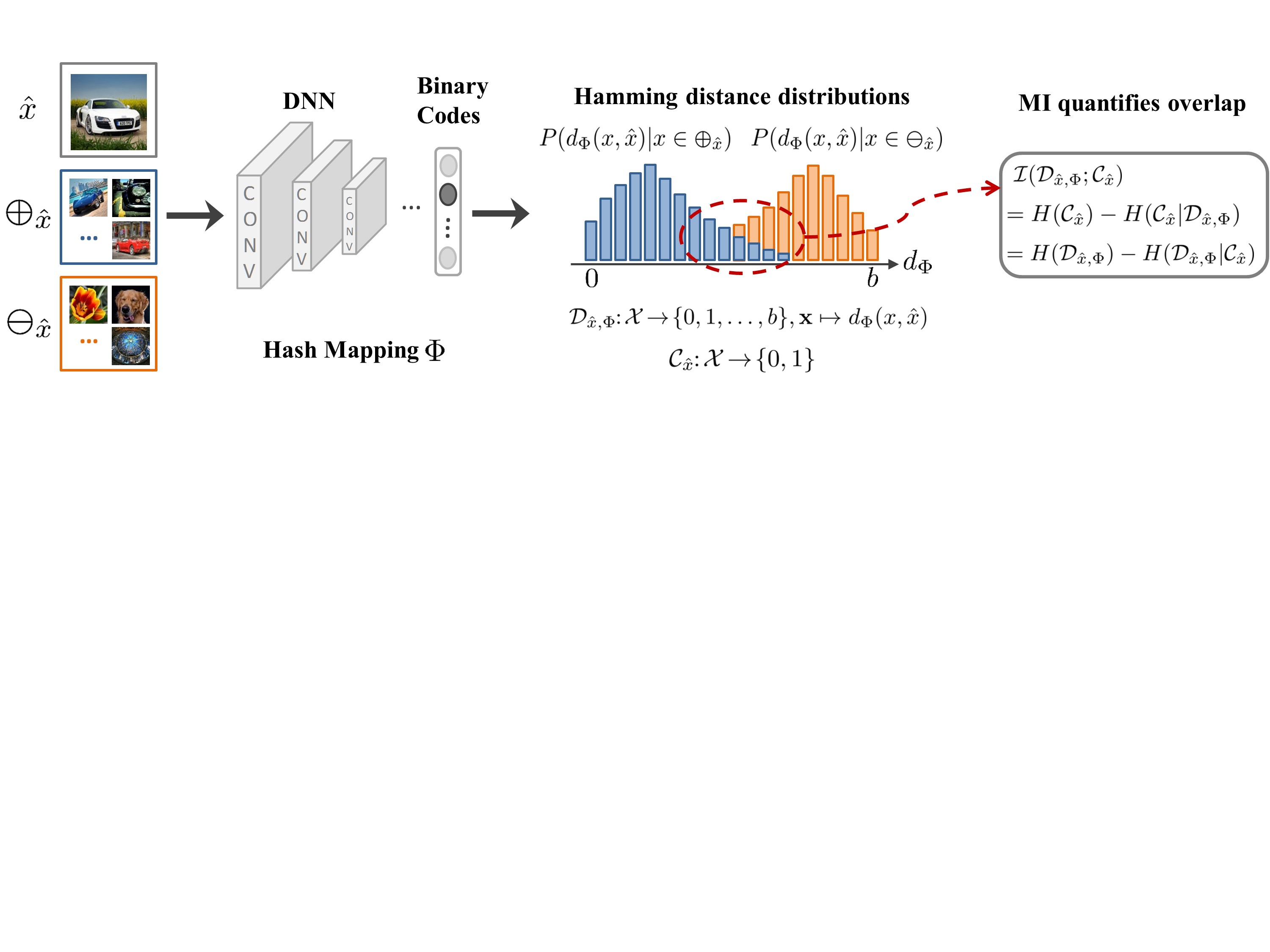}
\vspace{-1em}
\caption{Overview of the proposed hashing method. We use a deep neural network to compute $b$-bit binary codes  for: a (1) query image $\xhat$, (2) its neighbors in $\vp_\xhat$, and (3) its non-neighbors in $\vm_\xhat$. The binary codes are obtained via thresholding the activations in the last layer of the neural network. Computing hamming distances between the binary code of the query and the binary codes of neighbors and non-neighbors yields two distributions of Hamming distances. The information-theoretic quantity, Mutual Information, can be used to capture the separability between these two distributions, which gives a good quality indicator and learning objective. 
}
\label{fig:mi}
\end{figure*} 

\section{Hashing with Mutual Information}
\label{sec:formulation}
\subsection{Preliminaries}
Let $\mathcal{X}\subset\mathbb{R}^N$ be the feature space and $\mathcal{H}^b$ be the $\nbits$-dimensional Hamming space, \ie, $\mathcal{H}^b=\{-1,1\}^b$.
The goal of hashing is to learn an {embedding} function $\Phi:\mathcal{X}\to \mathcal{H}^\nbits$, which induces a Hamming distance $d_\Phi:\mathcal{X}\times\mathcal{X}\to\{0,1,\ldots,\nbits\}$ that is equal to the number of bit differences between embedded vectors.

We consider a supervised learning setup. For some example $\xhat\in\mathcal{X}$,  we assume that we have access to a set $\vp_\xhat\subset\mathcal{X}$ containing examples that are labeled as similar to $\xhat$ (neighbors), and  a set $\vm_\xhat$ of dissimilar examples (non-neighbors).
We also assume that this similar/dissimilar relationship is symmetric: 
 $x\in\vp_\xhat\Leftrightarrow \xhat\in\vp_x$, and $x\in\vm_\xhat\Leftrightarrow \xhat\in\vm_x$.
We call $\xhat$ an anchor example, and refer to 
$(\vp_\xhat,\vm_\xhat)$
as its \emph{neighborhood structure}. 
Then, we can cast the problem of learning the Hamming embedding $\Phi$ as one of preserving the neighborhood structure:
the neighbors of $\xhat$ should be mapped to the close vicinity of $\xhat$ in the Hamming space, while the non-neighbors should be mapped farther away.
Ideally, we would like to satisfy the following constraint: 
\begin{align}
d_\Phi(\xhat,x_p)<d_\Phi(\xhat,x_n),~~\forall x_p\in\vp_\xhat,\forall x_n\in\vm_\xhat.
\label{eq:ns_preserve}
\end{align}
If the learned $\Phi$ successfully satisfies this constraint, then the neighborhood structure of $\xhat$ can be exactly recovered by thresholding the Hamming distance $d_\Phi(\xhat,\cdot)$.
Generally, the neighborhood structure can be constructed by repeatedly querying a pairwise similarity oracle 
$S$, \eg $S(x,\xhat)>0$ iff $x\in\vp_\xhat$, and $S(x,\xhat)<0$ iff $x\in\vm_\xhat$. 
In practice, such an oracle can be derived from agreement/disagreement of class labels, or from thresholding a distance metric (\eg Euclidean distance) in the original feature space $\mathcal{X}$.
We give concrete examples in Section~\ref{sec:exp}.

In this work, we parameterize the functional embeddings using deep neural networks (DNNs), as DNNs have recently shown to have superior learning capabilities when coupled with appropriate hardware acceleration. 
Also, in order to take advantage of end-to-end training by backpropagation,
we use gradient-based optimization, and adopt an equivalent formulation of the Hamming distance that is amenable to differentiation:
\begin{align}
 d_\Phi(x,x') & = \frac{1}{2}\left(\nbits-\Phi(x)^\top\Phi(x')\right),\label{eq:hamming_dist}\\
\Phi(x) & = (\phi_1(x),\ldots,\phi_\nbits(x)),\\
\phi_i(x) & = \text{sgn}(f_i(x; w))\in\{-1,+1\}, \forall i,
\label{eq:hashfunc}
\end{align}
where $f_i,\forall i$ are the activations produced by a feed-forward neural network, with learnable parameters $w$.

\subsection{Minimizing Neighborhood Ambiguity}
We now discuss a formulation for learning the Hamming embedding $\Phi$.
As mentioned above, given $\xhat$ and its neighborhood structure, we would like to satisfy Equation~\ref{eq:ns_preserve} as much as possible, or, minimize the amount of violation.
Indeed, many existing supervised hashing formulations are based on the idea of minimizing violations.
For instance, affinity matching methods, mentioned in Section~\ref{sec:introduction}, typically enforce the following constraints through their loss functions:
\begin{align}
d_\Phi(\xhat,x_p)<t_1, d_\Phi(\xhat,x_n)>t_2, ~\forall x_p\in\vp_\xhat,\forall x_n\in\vm_\xhat,
\end{align}
where $t_1\leq t_2$ are threshold parameters.
This indirectly enforces Equation~\ref{eq:ns_preserve} by constraining the absolute values of individual Hamming distances.
Alternatively, local ranking methods based on triplet supervision encourage  the following:
\begin{align}
d_\Phi(\xhat,x_p)+\eta\leq d_\Phi(\xhat,x_n), ~ \forall x_p\in\vp_\xhat,\forall x_n\in\vm_\xhat,
\end{align}
where $\eta$ is a margin parameter.
We note that both formulations are inflexible, since the same threshold or margin parameters are applied for all anchors $\xhat$. 
Also, it is often observed in practice that these parameters  are nontrivial to tune.

Instead, we propose a novel formulation based on the idea of minimizing \emph{neighborhood ambiguity}, which is more directly related to the quality of nearest neighbor retrieval.
We say that $\Phi$ induces neighborhood ambiguity if the mapped image of some $x_n\in\vm_\xhat$ is closer to that of $\xhat$ than some $x_p\in\vp_\xhat$ in the Hamming space. 
When this happens, it is no longer possible to exactly recover the neighborhood structure by thresholding $d_\Phi$.
Consequently, when $\Phi$ is used to perform retrieval, the retrieved ``nearest neighbors'' of $\xhat$ would be contaminated by non-neighbors.
Therefore, we conclude that a high-quality embedding should minimize neighborhood ambiguity.

To concretely formulate the idea, we define 
random variable $\mathcal{D}_{{\hat{x}},\Phi}\!:\!\mathcal{X}\!\to\!\{0,1,\ldots,b\}, \x\mapsto d_\Phi({x}, {\hat{x}})$, and let $\mathcal{C}_{\hat{{x}}}\!:\!\mathcal{X}\!\to\!\{0,1\}$  be the membership indicator for the set $\vp_\xhat$.
Then, we naturally have two conditional distributions of the Hamming distance: 
$P(\D_{\xhat,\Phi}|\C_\xhat=1)$ and $P(\D_{\xhat,\Phi}|\C_\xhat=0)$. 
Note that the constraint in Equation~\ref{eq:ns_preserve} can be re-expressed as having no overlap between these two conditional distributions, and that minimizing neighborhood ambiguity amounts to minimizing the overlap.
Please see Figure~\ref{fig:mi} for an illustration.

We use the \emph{mutual information} between random variables $\mathcal{D}_{{\hat{x}},\Phi}$ and $\mathcal{C}_{{\hat{x}}}$ to capture the amount of overlap between conditional Hamming distance distributions. The mutual information is defined as
\begin{align}
\mathcal{I}(\mathcal{D}_{{\hat{x}},\Phi}; \mathcal{C}_{{\hat{x}}}) 
& = H(\mathcal{C}_{{\hat{x}}}) - H(\mathcal{C}_{{\hat{x}}} | \mathcal{D}_{{\hat{x}},\Phi})\\
& = H(\D_{\xhat,\Phi})-H(\D_{\xhat,\Phi}|\C_{\xhat})
\label{eq:mi}
\end{align}
where $H$ denotes (conditional) entropy.
In the following, for brevity we will drop subscripts $\Phi$ and $\hat{{x}}$, and denote the two conditional distributions and the marginal $P(\D_{\xhat,\Phi})$ as $p_\D^+$, $p_\D^-$, and $p_\D$, respectively.

By definition, $\mathcal{I}(\D;\C)$  measures the decrease in uncertainty in the neighborhood information $\mathcal{C}$ when observing the Hamming distances $\mathcal{D}$.
If $\mathcal{I}(\D;\C)$ attains a high value, which means $\mathcal{C}$ can be determined with low uncertainty by observing $\mathcal{D}$, then $\Phi$ must have achieved good separation (\textit{i.e.} low overlap) between $p_{\mathcal{D}}^+$ and $p_{\mathcal{D}}^-$.
$\mathcal{I}$ is maximized when there is no overlap, and minimized when $p_{\mathcal{D}}^+$ and $p_{\mathcal{D}}^-$ are exactly identical. 
As $H(\mathcal{C}_{{\hat{x}}})$ is typically constant, maximizing mutual information corresponds to minimizing the conditional entropy $ H(\mathcal{C}_{{\hat{x}}} | \mathcal{D}_{{\hat{x}},\Phi})$. 
Note that this conditional entropy directly corresponds to the neighborhood ambiguity when $\mathcal{D}$ is observed. 
Mutual information is also related to the \textit{Kullback-Leibler} divergence measure $D_{KL}$, specifically as 
\begin{align}
\mathcal{I}(\mathcal{D}; \mathcal{C}) = \mathbb{E}_{\mathcal{D}} \llbracket D_{KL}(P(\mathcal{C} |\mathcal{D}) ||P(\mathcal{C})) \rrbracket,
\end{align} 
corresponding to the expected divergence between the distributions $P(\mathcal{C} |\mathcal{D})$ and $P(\mathcal{C})$.
Intuitively, if $\mathcal{D}$ were to be informative, these two Bernoulli distributions should differ. 
Indeed, maximizing the $D_{KL}$ divergence maximizes the difference of the two distributions.

Next, for any hash mapping $\Phi$, we can integrate $\mathcal{I}$ over the feature space to give a quality measure:
\begin{equation}
\mathcal{O}(\Phi) = \int_{\mathcal{X}} \mathcal{I}({\mathcal{D}_{{\hat{x}},\Phi}};C_{{\hat{x}}})p({\hat{x}})d{\hat{x}}.
\label{mi_batch_criteria}
\end{equation} 
An appealing property of this mutual information objective is that it is \emph{parameter-free}: the objective encourages distributions $p_\D^+$ and $p_\D^-$ to be separated, but does not include parameters dictating the distance threshold at which separation occurs, or the absolute amount of separation.
The absence of such fixed parameters also increases flexibility, since the separation could occur at different distance thresholds depending on the anchor $\xhat$.

\section{Optimizing Mutual Information}
\label{sec:optim}

Having shown that mutual information is a suitable measure of hashing quality, we consider its use as a learning objective. 

Clearly, optimizing $\mathcal{O}(\Phi)$, as defined in Equation~\ref{mi_batch_criteria}, is intractable.
As is usually the case in supervised learning, we optimize the parameters of $\Phi$ over a finite training set $\mathcal{T}$ of i.i.d. samples from $p(\xhat)$. 
Our learning problem is then formulated as
\begin{align}
\max_{\Phi} ~ \frac{1}{|\mathcal{T}|}\sum_{x\in\mathcal{T}}\mathcal{I}(\D_{x,\Phi}; \C_x).
\label{eq:mi_obj}
\end{align}
It is worth noting that for each $x\in\mathcal{T}$, elements of $\vp_x$ and $\vm_x$ are now restricted to be within $\mathcal{T}$.
Inspired by recent advances in stochastic optimization, we will use stochastic gradient descent to solve the above problem.

We start by deriving the gradients of $\mathcal{I}$ with respect to the output of the hash mapping, $\Phi(x)$.
First, note that with $b$-bit Hamming distances, the discrete distributions $p_\D^+$ and $p_\D^-$ can be modeled using normalized histograms over $\{0,\ldots,b\}$.
Specifically, 
let $p_{\D,l}^+$ be the $l$-th element of $p_{\D}^+$, 
which is estimated by performing hard assignments on Hamming distances into histogram bins:
\begin{equation}
p_{\mathcal{D},l}^+ = \frac{1}{|\vp_\xhat|} \sum_{x\in\vp_\xhat} \boldsymbol{1}[d_\Phi(\xhat,x)=l], ~ l=0,\ldots,b,
\label{eq:binning}
\end{equation}
where $\boldsymbol{1}[\cdot]$ denotes the binary indicator.

The mutual information $\mathcal{I}$ is continuously differentiable, and  using the chain rule we can write
\begin{align}
\frac{\partial \mathcal{I}}{\partial \Phi(x)}=\sum_{l=0}^b\left[\frac{\partial \mathcal{I}}{\partial p^+_{\D,l}}\frac{\partial p^+_{\D,l}}{\partial \Phi(x)}
+ \frac{\partial \mathcal{I}}{\partial p^-_{\D,l}}\frac{\partial p^-_{\D,l}}{\partial \Phi(x)}\right].
\label{eq:diff_mi}
\end{align}
Due to symmetry, we next only focus on terms involving $p_{\D}^+$.
Let $p^+$ and $p^-$ be shorthands for the priors $P(\C=1)$ and $P(\C=0)$. 
For $l=0,\ldots,b$, we have
\begin{align}
\frac{\partial \mathcal{I}}{\partial p^+_{\D,l}} & =
-\frac{\partial H(\D|\C)}{\partial p^+_{\D,l}}+\frac{\partial H(\D)}{\partial p^+_{\D,l}}\\
& = p^+(\log p^+_{\D,l}+1)-(\log p_{\D,l}+1)\frac{\partial p_{\D,l}}{\partial p^+_{\D,l}}\\
& = p^+(\log p^+_{\D,l}-\log p_{\D,l}).
\label{eq:mi_deriv}
\end{align}
Note that for Equation~\ref{eq:mi_deriv}, we used the fact that 
\begin{align}
p_{\D,l}=p^+p^+_{\D,l}+p^-p^-_{\D,l}.
\end{align}

\subsection{Continuous Relaxation}

To complete the chain rule, we need to further derive the term ${\partial p_{\D,l}^+}/{\partial \Phi(x)}$ in Equation~\ref{eq:diff_mi}.
However, the hash mapping $\Phi$ is discrete by nature, precluding the use of continuous optimization.
While it is possible to maintain such constraints and resort to discrete optimization, 
the resulting optimization problems are NP-hard.

Instead, in order to apply gradient-based continuous optimization, we take the relaxation approach to sidestep the NP-hard problems. Correspondingly, we need to perform a continuously differentiable relaxation to $\Phi$.
Recall from Equation~\ref{eq:hashfunc} that each element in $\Phi$ is obtained by thresholding neural network activations with the sign function. 
We relax $\Phi$ into a real-valued vector $\hat{\Phi}$ by adopting a standard technique in hashing \cite{cakir2015adaptive,Li_IJCAI2016,SHK}, where the  discontinuous sign function is approximated with the sigmoid function $\sigma:\mathbb{R}\to(0,1)$: 
\begin{align}
\hat{\Phi}(x)&=(\hat{\phi}_1(x),\ldots,\hat{\phi}_b(x)),\\
\hat{\phi}_i(x)&=2\sigma(\gamma f_i(x;w))-1\in (-1,1).
\label{eq:sigmoid}
\end{align}

We include a tuning parameter $\gamma$, used to control the ``{steepness}'' of the sigmoid approximation.
Typically, we choose $\gamma \geq 1$ so as to reduce the error introduced by the continuous relaxation.
Although with $\gamma\rightarrow\infty$ the sigmoid approximation approaches the sign function, a large $\gamma$ can make gradients vanish due to the saturation of the sigmoid function. 
We empirically evaluate the choice of $\gamma$ with an ablation study in Section~\ref{sec:exp}, and find that \mihash~is quite robust with respect to the continuous relaxation.
Other alternative relaxation strategies include using a quantization error term \cite{Li_IJCAI2016,Wang_ACCV2016} and applying the continuation method \cite{hashnet}. 

With the continuous relaxation in place, we move on to the partial differentiation of  $p_\D^+$ and $p_\D^-$ with respect to $\hat{\Phi}(x)$.
As mentioned before, these discrete distributions can be estimated via histogram binning (Equation~\ref{eq:binning});  however, histogram binning is a non-differentiable operation, due to the use of the binary indicator function.
In the following, we describe a differentiable approximation to the discrete histogram binning process, thereby enabling end-to-end backpropagation.

\subsection{End-to-End Optimization}
Without the continuous relaxation, Equation~\ref{eq:binning} performs histogram binning by assigning $d_\Phi(\xhat,x)$, which is an integer, into a specific bin. 
With the continuous relaxation developed above, we note that $d_\Phi$ in is no longer integer-valued, but is also continuously relaxed into
\begin{align}
\hat{d}_\Phi(\xhat,x)=\frac{b-\hat{\Phi}(\xhat)^\top\hat{\Phi}(x)}{2}.
\label{eq:hamming_dist_relaxed}
\end{align}

When $d_\Phi$ is relaxed into $\hat{d}_\Phi$, we need to replace the hard assignment with soft assignment. 
The key is to approximate the binary indicator $\boldsymbol{1}[\cdot]$ with a differentiable function.
For this purpose, we employ a technique from \cite{ustinova2016learning}.
Specifically, the binary indicator is replaced by a triangular kernel function $\delta$ with slope parameter $\Delta>0$, centered on the histogram bin center, which linearly interpolates the real-valued $\hat{d}_\Phi(\xhat,x)$ into the $l$-th bin:
\begin{align}
\delta(d,l) =
\max\left\{0, 1-\frac{|d - l|}{\Delta}\right\}.
\end{align}
It is easy to see that this triangular approximation approaches the original binary indicator as $\Delta\to 0$. 
Also, this soft assignment admits simple subgradients:
\begin{equation}
  \delta_l'(d)\overset{\Delta}{=}\frac{\partial \delta(d,l)}{\partial d} =
  \begin{cases}
  1/\Delta, & d \in [l-\Delta, l], \\
  -1/\Delta, & d \in [l, l+\Delta], \\
  0, & \text{otherwise.}
  \end{cases}
  \label{eq:delta-dh}
\end{equation}

We are now ready to tackle the term $\partial p^+_{\D,l}/\partial\hat{\Phi}(x)$. 
From the definition of $p_{\D,l}^+$ in Equation~\ref{eq:binning}, we have, for $x=\xhat$:
\begin{align}
\frac{\partial p_{\mathcal{D},l}^+}{\partial \hat{\Phi}(\xhat)}
&= \frac{1}{|\vp_\xhat|} \sum_{x \in \vp_\xhat} \frac{\partial \delta(\hat{d}_\Phi(\xhat,x),l)}{\partial \hat{\Phi}(\xhat)}\\
& = \frac{1}{|\vp_\xhat|} \sum_{x \in \vp_\xhat} \frac{\partial \delta(\hat{d}_\Phi(\xhat,x),l)}{\partial \hat{d}_\Phi(\xhat,x)}\frac{\partial \hat{d}_\Phi(\xhat,x)}{\partial \hat{\Phi}(\xhat)}\\
&= \frac{-1}{2|\vp_\xhat|} \sum_{x \in \vp_\xhat}  \delta_l'(\hat{d}_\Phi(\xhat,x)) \hat{\Phi}(x).
\label{eq:condprob-phi}
\end{align}

For the last step, we used the definition of $\hat{d}_\Phi$ in Equation~\ref{eq:hamming_dist_relaxed}.
Next, for $x\neq \xhat$:
\begin{align}
\frac{\partial p_{\mathcal{D},l}^+}{\partial \hat{\Phi}(x)}
&= \frac{1}{|\vp_\xhat|} \boldsymbol{1}[x \in \vp_\xhat] 
\frac{\partial \delta(\hat{d}_\Phi(\xhat,x),l)}{\partial \hat{d}_\Phi(\xhat,x)}
\frac{\partial \hat{d}_\Phi(\xhat,x)}{\partial \hat{\Phi}(x)}\\
&= \frac{-1}{2|\vp_\xhat|} \boldsymbol{1}[x \in \vp_\xhat]  \delta_l'(\hat{d}_\Phi(\xhat,x)) \hat{\Phi}(\xhat).
\label{eq:condprob-phi}
\end{align}

Lastly, to back-propagate gradients to $\hat{\Phi}$'s {inputs}, and ultimately to the parameters of the underlying deep neural network, we only need to further differentiate the sigmoid approximation employed in $\hat{\Phi}$ (Equation~\ref{eq:sigmoid}). 
The derivative of the sigmoid function has a closed form expression, and is omitted here.

\subsection{Efficient Minibatch Backpropagation}
So far, our derivations of mutual information and its gradients have assumed a single anchor example $\xhat$. 
In information retrieval terminology, the current derivations are for a single query and a fixed database. 
However, the optimization objective in Equation~\ref{eq:mi_obj} is the average of mutual information values over all anchors in a finite training set $\mathcal{T}$. 
We now address this mismatch.

We face two challenges when working with a (potentially large) training set $\mathcal{T}$.
First, we need to perform the optimization in the stochastic/minibatch setting, since deep neural networks are typically trained by minibatch stochastic gradient descent (SGD), where it can be infeasible to access the entire database all at once.
The second challenge is that, differently from traditional information retrieval, in many computer vision tasks (\eg image retrieval), there is usually no clear split of a given training set into a set of queries and a database. 
This is due to the symmetry that an image can either be a query or a database item. 
Consequently, even if we {were to} create such a split, it can be arbitrary and does not fully utilize available supervision. 

Here, we describe a way to efficiently utilize all the available supervision during minibatch SGD training, simultaneously addressing both challenges.
Our reasoning is that, within a minibatch with $M$ examples, a retrieval problem can be defined by retrieving one example (the query) against the other $M-1$ examples (the database).
Further, considering the symmetry mentioned above, retrieval can be repeated $M$ times, each time using a different example as the query.
Then, the overall objective value for the minibatch is the average over the $M$ individual retrieval problems. 
This way, the available supervision in the minibatch is utilized maximally.
As we shall see next, the backpropagation in this case can be efficiently implemented using matrix multiplications.

Now consider a minibatch of size $M$, $\mathcal{B}=\{x_1,\ldots,x_M\}$. 
Since we only operate within the minibatch, for each example $x_i\in \mathcal{B},1\leq i\leq M$, when used as the query, its neighborhood structure is now defined within $\mathcal{B}$: 
we take $\vp_i= \vp_{x_i}\cup \mathcal{B}$, and $\vm_i= \vm_{x_i}\cup \mathcal{B}$.
Also, let $\mathcal{I}_i$ be a shorthand for $\mathcal{I}(\mathcal{D}_{x_i,\Phi},C_{x_i})$.
We group the relaxed hash mapping output for the entire minibatch into the following $b\times M$ matrix,
\begin{align}
\boldsymbol{\hat{\Phi}} =  \left[
\hat{\Phi}(x_1) ~ \hat{\Phi}(x_2) ~ \cdots ~ \hat{\Phi}(x_M)
\right]\in\mathbb{R}^{b\times M}.
\end{align}
Similar to Equation~\ref{eq:diff_mi}, we can write the Jacobian matrix of the minibatch objective $\mathcal{O}_{\mathcal{B}}$ with respect to $\bigPhi$ as

\begin{align}
\frac{\partial \mathcal{O}_{\mathcal{B}}}{\partial \bigPhi} 
& = \frac{1}{M} \sum_{i=1}^M\frac{\partial \mathcal{I}_i}{\partial \bigPhi}  \\
& = \frac{1}{M}\left[\sum_{i=1}^M\sum_{l=0}^b
\frac{\partial \mathcal{I}_i}{\partial p_{i,l}^+} \frac{\partial p_{i,l}^+}{\partial \bigPhi }
+ \sum_{i=1}^M\sum_{l=0}^b
\frac{\partial \mathcal{I}_i}{\partial p_{i,l}^-} \frac{\partial p_{i,l}^-}{\partial \bigPhi } 
\right]
\label{eq:grad_minibatch}
\end{align}
where $p_{i,l}^+$ ($p_{i,l}^-$) denotes the $l$-th element of $p_{\D}^+$ ($p_{\D}^-$) when the query is $x_i$.

Again, we have covered the derivation of the partial derivative $\partial \mathcal{I}_i/\partial p_{i,l}^+$,
and now the main issue is evaluating the Jacobian $\partial p_{i,l}^+/\partial \bigPhi$. 
We do so by examining each column of the Jacobian. 
First, for $\forall j\neq i$,
\begin{align}
\frac{\partial p_{i,l}^+}{\partial \hat{\Phi}(x_j)} 
& = \frac{\partial p_{i,l}^+}{\partial \hat{d}_\Phi(x_i,x_j)} \frac{\partial \hat{d}_\Phi(x_i,x_j)}{\partial \hat{\Phi}(x_j)}\\
& = \frac{\boldsymbol{1}[x_j\in\vp_i]}{|\vp_i|}\delta_l'(\hat{d}_\Phi(x_i,x_j))\frac{-\hat{\Phi}(x_i)}{2}\\
& \overset{\Delta}{=} \frac{\beta_l^+(i,j)}{N_i^+}\frac{-\hat{\Phi}(x_i)}{2},
\end{align}
where we have made the following substitutions:
\begin{align}
N_i^+ & = |\vp_i|, \\
\beta_l^+(i,j) & = \boldsymbol{1}[x_j\in\vp_i] \delta_l'(\hat{d}_\Phi(x_i,x_j)).
\label{eq:beta}
\end{align}

Next, for $j=i$,
\begin{align}
\frac{\partial p_{i,l}^+}{\partial \hat{\Phi}(x_i)} & = 
\sum_{k\neq i}\frac{\partial p_{i,l}^+}{\partial \hat{d}_\Phi(x_i,x_k)} \frac{\partial \hat{d}_\Phi(x_i,x_k)}{\partial \hat{\Phi}(x_i)}  \\
& = \sum_{k\neq i} \frac{\beta_l^+(i,k)}{N_i^+}\frac{-\hat{\Phi}(x_k)}{2} .
\end{align}

By defining that $\beta_l^+(i,i)\equiv 0,\forall i$,
we can  further unify these two cases as:
\begin{align}
\frac{\partial p_{i,l}^+}{\partial \hat{\Phi}(x_j)} = 
-\frac{\beta_l^+(i,j)\hat{\Phi}(x_i)}{2N_i^+}
- \boldsymbol{1}[j=i] \sum_{k=1}^M \frac{\beta_l^+(i,k)\hat{\Phi}(x_k)}{2N_i^+}.
\end{align}

Having derived all the columns, we now write down the matrix form of $\partial p_{i,l}^+/\partial \bigPhi$. 
First, we define $\boldsymbol{\beta}^+_{l,i}=(\beta_l^+(i,1),\ldots,\beta_l^+(i,M))\in\mathbb{R}^M$, and let $\boldsymbol{e}_i$ be the $i$-th standard basis vector in $\mathbb{R}^M$ (\ie\!\!, the $i$-th element is 1 and other elements are 0).
The matrix form can be compactly written as
\begin{align}
\frac{\partial p_{i,l}^+}{\partial \bigPhi} & = 
-\frac{\hat{\Phi}(x_i)(\boldsymbol{\beta}_{l,i}^+)^\top}{2N_i^+}
-\left[\sum_{k=1}^M\frac{\beta_l^+(i,k)\hat{\Phi}(x_k)}{2N_i^+}\right]\boldsymbol{e}_i^\top
\\ & = 
-\frac{1}{2N_i^+}\left[\hat{\Phi}(x_i)(\boldsymbol{\beta}_{l,i}^+)^\top
+ \bigPhi \boldsymbol{\beta}^+_{l,i}\boldsymbol{e}_i^\top\right].
\end{align}

We will next complete Equation~\ref{eq:grad_minibatch}.  First, we define a shorthand, which can be easily evaluated using the result in Equation~\ref{eq:mi_deriv}:
\begin{align}
\alpha_{l,i}^+=\frac{1}{N_i^+}\frac{\partial \mathcal{I}_i}{\partial p_{i,l}^+}.
\end{align}
Using symmetry, we only consider the first term involving $p_{i}^+$  in Equation~\ref{eq:grad_minibatch}, and we omit the $1/M$ scaling factor for now:
\begin{align}
& \sum_{i=1}^M\sum_{l=0}^b
\frac{\partial \mathcal{I}_i}{\partial p_{i,l}^+} \frac{\partial p_{i,l}^+}{\partial \bigPhi }\\
 = &\sum_{l=0}^b\sum_{i=1}^M -\frac{1}{2N_i^+}\frac{\partial \mathcal{I}_i}{\partial p_{i,l}^+}
\left[ \hat{\Phi}(x_i)(\boldsymbol{\beta}_{l,i}^+)^\top
+ \bigPhi \boldsymbol{\beta}^+_{l,i}\boldsymbol{e}_i^\top \right]\\
 = &-\frac{1}{2}\sum_{l=0}^b\alpha_{l,i}^+\left[
\sum_{i=1}^M\hat{\Phi}(x_i)(\boldsymbol{\beta}_{l,i}^+)^\top
+ \bigPhi \sum_{i=1}^M\boldsymbol{\beta}_{l,i}^+\boldsymbol{e}_i^\top
\right]\\
 = &-\frac{1}{2}\sum_{l=0}^b\left[
\sum_{i=1}^M\alpha_{l,i}^+\hat{\Phi}(x_i)(\boldsymbol{\beta}_{l,i}^+)^\top
+ \bigPhi \sum_{i=1}^M\alpha_{l,i}^+\boldsymbol{\beta}_{l,i}^+\boldsymbol{e}_i^\top \right].
\label{eq:intermediate}
\end{align}
Define 
\begin{align}
A_l^+ & =\mathrm{diag} (\alpha_{l,1}^+,\ldots,\alpha_{l,M}^+)\in\mathbb{R}^{M\times M},\label{eq:A} \\
B_l^+ & =\left[ \boldsymbol{\beta}_{l,1}^+ ~ \cdots ~ \boldsymbol{\beta}_{l,M}^+\right]
\in \mathbb{R}^{M\times M},
\label{eq:B}
\end{align}
then we can simplify Equation~\ref{eq:intermediate} as
\begin{align}
& -\frac{1}{2}\sum_{l=0}^b\left[ \bigPhi A_l^+ (B_l^+)^\top  +\bigPhi B_l^+ A_l^+ \right] \\
= &  -\frac{1}{2}\bigPhi \sum_{l=0}^b\left(  A_l^+B_l^+  + B_l^+ A_l^+ \right).
\end{align}
The last step is true since $B_l^+$ is symmetric: 
it can be seen from the definition of $\beta^+$ in Equation~\ref{eq:beta} that $\beta^+_l(i,j)=\beta^+_l(j,i)$, since both the neighbor relationship and the Hamming distance are symmetric.

Now, if we define $A_l^-$ and $B_l^-$ for the non-neighbor distance distribution $p_\D^-$, analogously as in Equations~\ref{eq:A} and \ref{eq:B} (details are very similar and omitted), then the full Jacobian matrix in Equation~\ref{eq:grad_minibatch} can be evaluated as
\begin{align}
-\frac{\bigPhi}{2M}\sum_{l=0}^b( A_l^+B_l^+  + B_l^+ A_l^+ + A_l^-B_l^-  + B_l^- A_l^-).
\label{eq:minibatch_bp}
\end{align}
Since only matrix multiplications and additions are involved, this operation can be implemented efficiently.
In particular, note that $A_l^+$ ($A_l^-$) is a diagonal matrix, therefore multiplying with $B_l^+$ ($B_l^-$) effectively scales its rows or columns, which has $O(M^2)$ time complexity, as opposed to general matrix multiplication which is $O(M^3)$.
We then conclude that the overall time complexity for computing Equation~\ref{eq:minibatch_bp} is $O(bM^2)$.

Recently, Triantafillou \etal \cite{fewshot} also propose  a minibatch-based learning formulation that is inspired by information retrieval, which attempts to maximize the utilization of supervision by treating each example in the minibatch as a query. 
However, we note that \cite{fewshot} tackles the problem of few-shot learning by learning real-valued embeddings, and it uses very different optimization machinery to approximately optimize mean Average Precision in a structured prediction framework.
Nevertheless, it would be interesting to explore the use of hashing and the mutual information objective for that problem in future work.

\section{Experiments}
\label{sec:exp}

\subsection{Datasets and Evaluation Setup}
We conduct experiments on widely used image retrieval benchmarks: CIFAR-10 \cite{krizhevsky2009learning}, NUSWIDE \cite{nuswide}, 22K LabelMe \cite{russell2008labelme} and ImageNet100 \cite{deng2009imagenet}. 
Each dataset is split into a test set and retrieval set, and instances from the retrieval set are used in training.
We follow a standard information retrieval setup: at test time, queries from the test set are used to rank instances from the retrieval set using Hamming distances, and the performance metric is averaged over the queries.
\medskip
\begin{itemize}

\item \textbf{CIFAR-10} is a dataset for image classification and retrieval, containing 60K images from 10 different categories. We follow the setup of \cite{Lai_CVPR2015, Zhuang_CVPR2016, Li_IJCAI2016, Wang_ACCV2016}. This setup corresponds to two distinct partitions of the dataset. In the first case (\cifar{1}), we sample 500 images per category, resulting in 5,000 training examples to learn the hash mapping. The test set contains 100 images per category (1000 in total). The remaining images are then used to populate the hash table. In the second case (\cifar{2}), we sample 1000 images per category to construct the test set (10,000 in total). The remaining items are both used to learn the hash mapping and populate the hash table. Two images are considered neighbors if they belong to the same class. 

\medskip
\item \textbf{NUSWIDE} is a dataset containing 269K images from Flickr. Each image can be associated with multiple labels, corresponding with 81 ground truth concepts. For NUSWIDE experiments, following the setup in \cite{Lai_CVPR2015, Zhuang_CVPR2016, Li_IJCAI2016, Wang_ACCV2016}, we only consider images annotated with the 21 most frequent labels. In total, this corresponds to 195,834 images. The experimental setup also has two distinct partitionings: \nus{1} and \nus{2}. For both cases, a test set is constructed by randomly sampling 100 images per label (2,100 images in total). 
To learn the hash mapping, 500 images per label are randomly sampled in \nus{1} (10,500 in total). The remaining images are then used to populate the hash table. 
In the second case, \nus{2}, all the images excluding the test set are used in learning and populating the hash table. 
Following standard practice, two images are considered as neighbors if they share at least one label. 

\medskip
\item \textbf{22K LabelMe} consists of 22K images, each represented with a 512-dimensionality GIST descriptor. Following \cite{BRE, cakir2015adaptive}, we randomly partition the dataset into  a retrieval and a test set, consisting of 20K and 2K instances, respectively. A 5K subset of the retrieval set is used in learning the hash mapping. As this dataset is unsupervised, we use the Euclidean distance between GIST features in determining the neighborhood structure. Two examples that have a Euclidean distance below the $5\%$ distance percentile are considered neighbors.

\medskip
\item \textbf{ImageNet100} is a subset of ImageNet \cite{deng2009imagenet} containing 130K images from 100 classes. We follow the setup in \cite{hashnet}, and randomly sample 100 images per class for training. All images in the selected classes from the ILSVRC 2012 validation set are used as the test set. Two images are considered neighbors if they belong to the same class.

\end{itemize}

\afterpage{
\begin{table*}[htbp]
\small
\centering
{
\begin{tabular}{l|v|v|v|v|v|v|v|v} 
\hline
$\mathsf{VGG-F}$ & \multicolumn{4}{c|}{ $\mathsf{CIFAR-10}$ } & \multicolumn{4}{c}{ $\mathsf{NUSWIDE}$ } \rule{0pt}{1em} \\
\hline
\multirow{2}{*}{ \textbf{Method} } & \multicolumn{4}{c|}{$\mathsf{mAP}$} & \multicolumn{4}{c}{$\mathsf{mAP@5K}$} \\ 
\cline{2-9} 
    & 12 Bits & 24 Bits   & 32 Bits   & 48 Bits & 12 Bits & 24 Bits   & 32 Bits   & 48 Bits \\ 
\hline
SH \cite{SH} & 0.183 & 0.164 & 0.161 & 0.161 & 0.621& 0.616 & 0.615 & 0.612\\
ITQ \cite{ITQ} & 0.237 & 0.246 & 0.255 & 0.261 & 0.719 & 0.739 & 0.747 & 0.756 \\
SPLH \cite{Wang_SPLH_ICML2010} & 0.299 & 0.33 & 0.335 & 0.33 & 0.753 & 0.775 & 0.783 & 0.786 \\
SHK \cite{SHK} & 0.488 & 0.539 & 0.548 & 0.563 & 0.768 & 0.804& 0.815 & 0.824\\
SDH \cite{Shen_CVPR2015} & 0.478 & 0.557 & 0.584 & 0.592 & \textbf{0.780} & 0.804 & {0.816} & 0.824 \\
FastHash \cite{FastHash} & 0.553 & 0.607 & 0.619 & 0.636 & 0.779 & 0.807 & {0.816} & {0.825}\\
StructHash \cite{LinIJCV2016} &  {0.664} & {0.693}  & {0.691} & {0.700} &  0.748 & 0.772 & 0.790 & 0.801 \\
VDSH \cite{Ziming_VeryDeep_CVPR2016} & 0.538 & 0.541 & 0.545 & 0.548 & 0.769 & 0.796 & 0.803 & 0.807\\
DPSH \cite{Li_IJCAI2016} & {0.713} & 0.727 & 0.744 & 0.757 & \f{0.758} & \f{0.793} & \f{0.818} & \f{0.830}\\
DTSH \cite{Wang_ACCV2016} & 0.710 & {0.750} & {0.765} & {0.774} & {0.773} & \f{0.813} & \f{0.820} & \f{0.838}\\
\mihash & \f{\textbf{0.738}} & \textbf{0.775} & \textbf{\f{0.791}} & \textbf{0.816} & 0.773 & \f{\textbf{0.820}} & \f{\textbf{0.831}} & \textbf{0.843} \\
\hline
\end{tabular}
\caption{Results on CIFAR-10 and NUSWIDE datasets with \cifar{1} and \nus{1} partitionings. The underlying deep learning architecture is \vggf. \mihash~outperforms competing methods on CIFAR-10, and shows improvements, especially with lengthier codes, on NUSWIDE.
}
\label{table:set-1}

\vspace{1cm}
\begin{tabular}{l|v|v|v|v|v|v|v|v} 
\hline
$\mathsf{VGG-F}$ & \multicolumn{4}{c|}{ $\mathsf{CIFAR-10}$ } & \multicolumn{4}{c}{ $\mathsf{NUSWIDE}$ } \rule{0pt}{1em} \\
\hline
\multirow{2}{*}{ \textbf{Method} } & \multicolumn{4}{c|}{$\mathsf{mAP}$} & \multicolumn{4}{c}{$\mathsf{mAP@50K}$} \\ 
\cline{2-9} 
	& 16 Bits & 24 Bits   & 32 Bits   & 48 Bits & 16 Bits & 24 Bits & 32 Bits   & 48 Bits \\
\hline
DRSH \cite{Zhao_CVPR2015} & 0.608 & 0.611 & 0.617 & 0.618 & 0.609 & 0.618 & 0.621 & 0.631 \\
DRSCH \cite{Zhang_TIP2015} & 0.615 & 0.622 & 0.629 & 0.631 & 0.715 & 0.722 & 0.736 & 0.741 \\
DPSH \cite{Li_IJCAI2016} & 0.903 & 0.885 & 0.915 & 0.911 & 0.715 & 0.722 & 0.736 & 0.741 \\
DTSH \cite{Wang_ACCV2016} & {0.915} & {0.923} & {0.925} & {0.926} & {0.756} & 0.776 & {0.785} & {0.799} \\
\mihash & \textbf{\f{0.927}} & \textbf{\f{0.938}} & \textbf{\f{0.942}} & \textbf{0.943} & \f{\textbf{0.798}} & \f{\textbf{0.814}} & \f{\textbf{0.819}} & \f{\textbf{0.820}} \\
\hline
\end{tabular}
\caption{Results on CIFAR-10 and NUSWIDE datasets with \cifar{2} and \nus{2} partitionings (with \vggf~architecture). \mihash~achieves new state-of-the-art performance, consistently improving over competing methods.
} 
\label{table:set-2}

\vspace{1cm}
\begin{tabular}{l|v|v|v|v} 
\hline
$\mathsf{AlexNet}$ & \multicolumn{4}{c}{ $\mathsf{ImageNet100}$ } \rule{0pt}{1em} \\
\hline
\multirow{2}{*}{ \textbf{Method} } & \multicolumn{4}{c}{$\mathsf{mAP@1K}$} \\ 
\cline{2-5} 
    & 16 Bits & 32 Bits   & 48 Bits   & 64 Bits  \\
\hline
DTSH \cite{Wang_ACCV2016} & 0.458 & 0.566 & 0.611 & 0.644 \\
HashNet \cite{hashnet}  & 0.506 &  0.630 & 0.663 & 0.683 \\
\mihash & \textbf{0.569} & \textbf{0.661} & \textbf{0.685} & \textbf{0.694} \\

\hline
\end{tabular}
\caption{$\mathsf{mAP@1K}$ values on $\mathsf{ImageNet100}$ using \alexnet. \mihash~outperforms HashNet, a state-of-the-art deep hashing formulation using continuation methods \cite{hashnet}.
}
\label{table:imagenet}

\vspace{1cm}
\begin{tabular}{l|v|v|v|v} 
\hline
$\mathsf{GIST}$ &  \multicolumn{4}{c}{ $\mathsf{22K~LabelMe}$ } \rule{0pt}{1em} \\
\hline
\multirow{2}{*}{ \textbf{Method} } & \multicolumn{4}{c}{$\mathsf{mAP}$}  \\
\cline{2-5} & 16 Bits & 32 Bits & 48 Bits   & 64 Bits \\
\hline
DPSH \cite{Li_IJCAI2016}  & 0.295 & 0.346 & 0.391 & 0.427 \\
DTSH \cite{Wang_ACCV2016} & 0.304 & 0.342 & 0.361 & 0.378 \\
FastHash \cite{FastHash} & 0.324 & 0.394 & 0.433 & 0.456 \\ 
StructHash \cite{LinIJCV2016} & 0.369 & 0.474 & 0.538 & 0.582 \\
\mihash & \textbf{0.384} & \textbf{0.496} & \textbf{0.554} & \textbf{0.598} \\
\hline
\end{tabular}
\caption{$\mathsf{22K~LabelMe}$ results with \gist~features. \mihash~significantly improves over the state-of-the-art methods. 
}
\label{table:labelme}
}

\end{table*}

}
 
\medskip
As for performance metric, we use the standard mean Average Precision ($\mathsf{mAP}$), or its variants.
We compare \mihash~against both classical and recent state-of-the-art hashing methods. These methods include: Spectral Hashing (\textbf{SH}) \cite{SH}, Iterative Quantization (\textbf{ITQ}) \cite{ITQ}, Sequential Projection Learning for Hashing (\textbf{SPLH}) \cite{Wang_SPLH_ICML2010}, Supervised Hashing with Kernels (\textbf{SHK}) \cite{SHK}, Fast Supervised Hashing with Decision Trees (\textbf{FastHash}) \cite{FastHash}, Structured Hashing (\textbf{StructHash}) \cite{LinIJCV2016}, Supervised Discrete Hashing (\textbf{SDH}) \cite{Shen_CVPR2015}, Efficient Training of Very Deep Neural Networks (\textbf{VDSH}) \cite{Ziming_VeryDeep_CVPR2016}, Deep Supervised Hashing with Pairwise Labels (\textbf{DPSH}) \cite{Li_IJCAI2016}, Deep Supervised Hashing with Triplet Labels (\textbf{DTSH}) \cite{Wang_ACCV2016}, and Hashing by Continuation (\textbf{HashNet}) \cite{hashnet}.
These competing methods have been shown to outperform earlier and other works such as \cite{SSH, BRE, MLH, Xia_AAAI2014, Lai_CVPR2015, Zhao_CVPR2015}. 

We finetune deep Convolutional Neural Network models that are pretrained on the ImageNet dataset, by replacing the final softmax classification layer with a new fully-connected layer that produces the binary bits.
The new fully-connected layer is randomly initialized.
For CIFAR-10 and NUSWIDE experiments, we finetune a VGG-F \cite{Chatfield14} architecture, as in \cite{Li_IJCAI2016,Wang_ACCV2016}. For ImageNet100 experiments, following the protocol of HashNet \cite{hashnet}, we finetune the AlexNet \cite{krizhevsky2012imagenet} architecture, and scale down the learning rate for pretrained layers by a factor of 0.01, since the model is finetuned on the same dataset for a different task.
For \textit{non-deep} methods, we use the output of the penultimate layer ($fc7$) of both architectures as input features, which are 4096-dimensional.
For the 22K LabelMe benchmark, all methods learn shallow models on top of precomputed 512-dimensional GIST descriptors. 
For gradient-based hashing methods, this corresponds to learning a single fully connected layer. 

We use SGD with momentum $0.9$ and weight decay of $5\times 10^{-4}$, and reduce the learning rate periodically by a predetermined factor ($0.5$ in most cases), which is standard practice.
During training, the minibatches are randomly sampled from the training set.

\subsection{Results}
Table \ref{table:set-1} gives results for \cifar{1} and \nus{1} experimental settings in which $\mathsf{mAP}$ and $\mathsf{mAP@5K}$ ($\mathsf{mAP}$ evaluated on the top 5,000 retrievals) are reported for the CIFAR-10 and NUSWIDE datasets, respectively. Deep learning based hashing methods such as DPSH and DTSH outperform most non-deep hashing solutions. 
This is not surprising as the hash mapping is learned simultaneously with feature learning. Non-deep solutions such as FastHash and SDH also perform competitively, especially in NUSWIDE experiments.
Our proposed method, \mihash, surpasses all competing methods in the majority of the experiments. For example, with 32 and 48-bit binary embeddings \mihash~surpasses the nearest competitor, DTSH, by $3\%$-$4\%$ in CIFAR-10. 
For NUSWIDE, \mihash~achieves state-of-the-art performances in all experiments excluding with 12 bits.

The performance improvement of \mihash\ is much more significant in the \cifar{2} and \nus{2} settings, where more training data is available.
In these settings, a {VGG-F} network pretrained on ImageNet is again finetuned. 
Following standard practice, $\mathsf{mAP}$ and $\mathsf{mAP@50K}$ metrics are used to evaluate the retrieval performance. Table \ref{table:set-2} gives the results. 
As can be observed, in both CIFAR-10 and NUSWIDE, \mihash~achieves state-of-the-art results in nearly all code lengths. For instance, \mihash~consistently outperforms DTSH, the closest competitor, 
by a large margin in all embedding sizes. 

\begin{figure*}[t]
\centering
\includegraphics[height=12em, trim= 4em 0em 4em 0em]{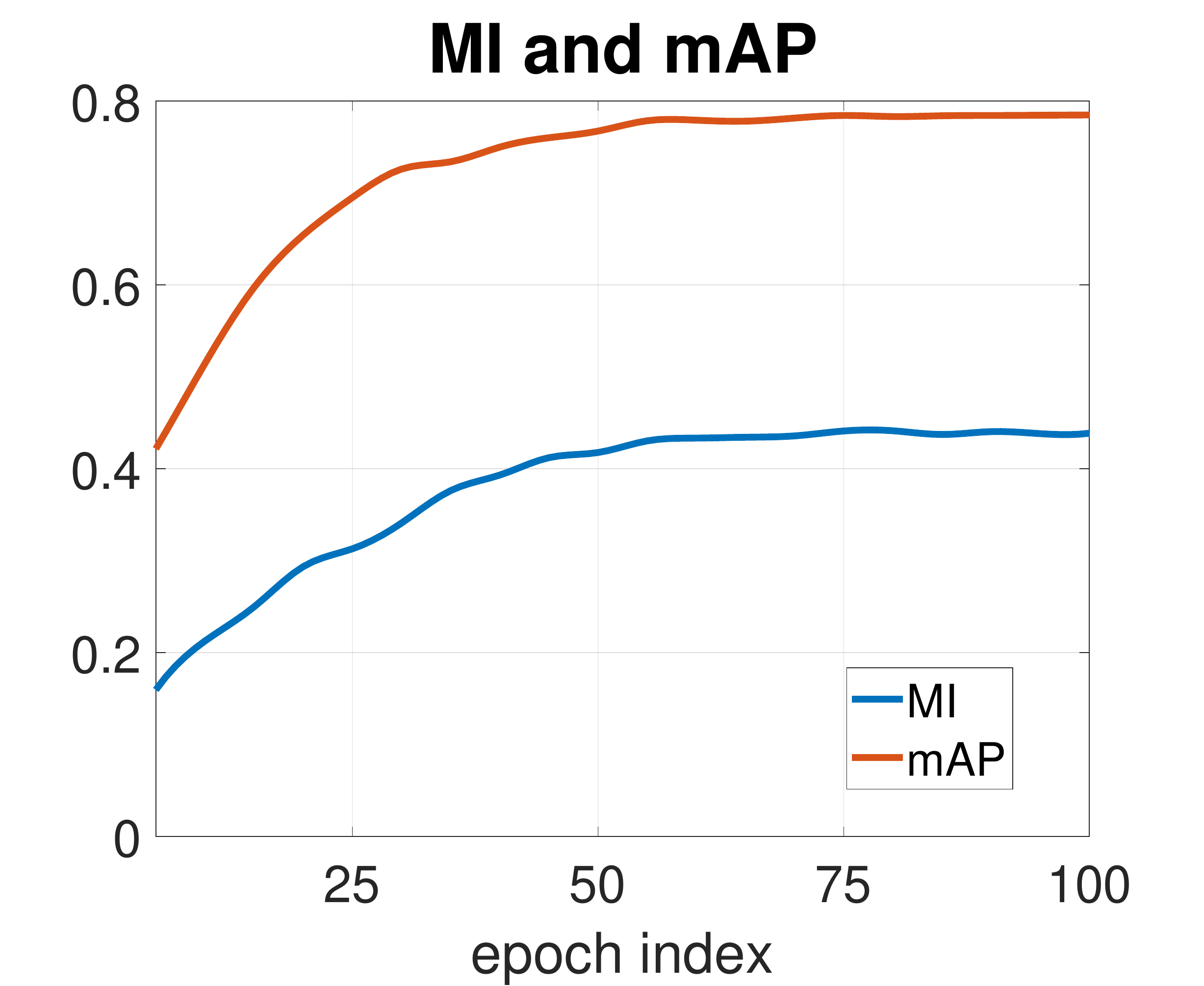}
~~~~\includegraphics[height=12em, trim= 4em 0em 4em 0em]{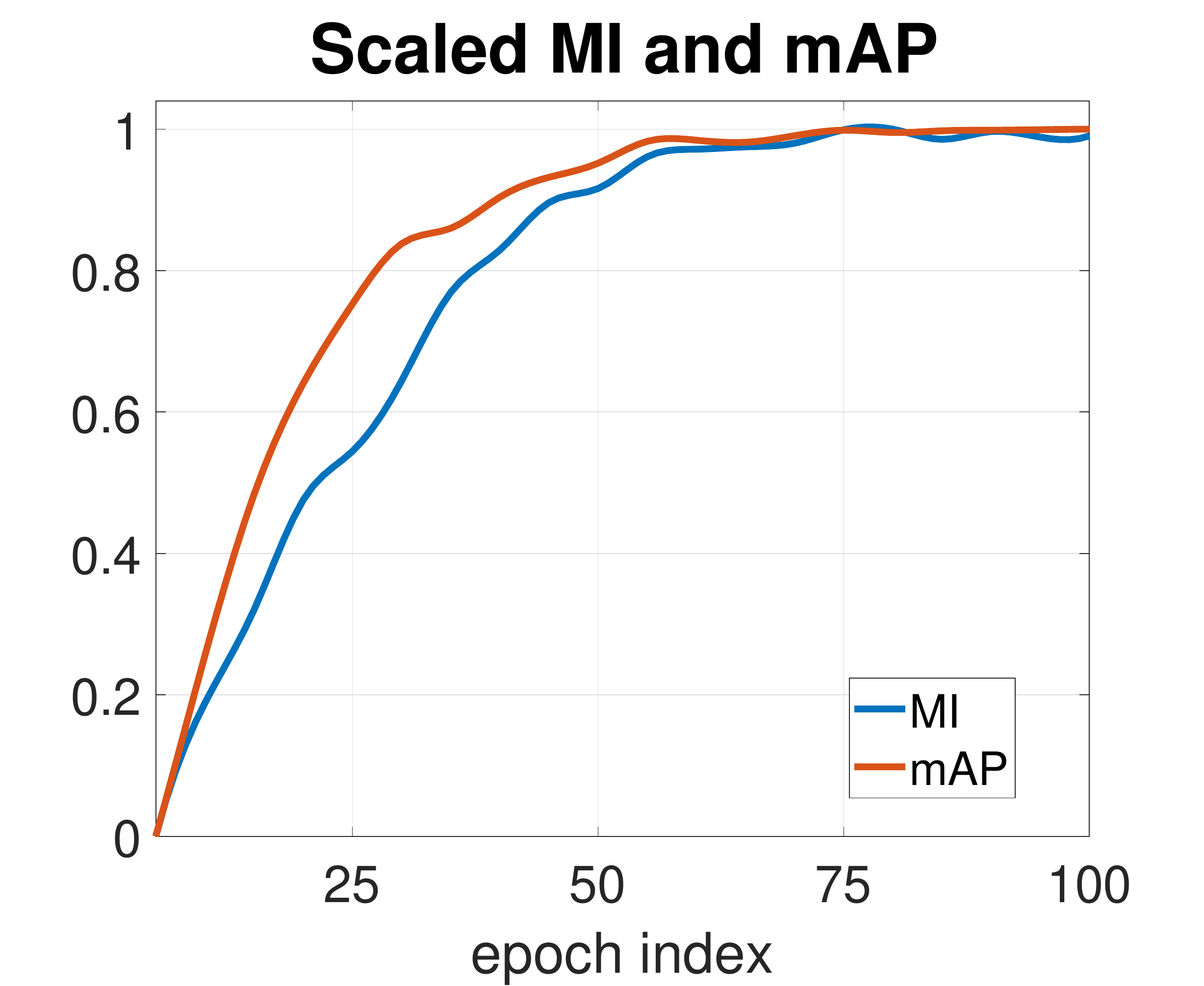}
~~~~~~\includegraphics[height=12em, trim= 4em 0em 4em 0em]{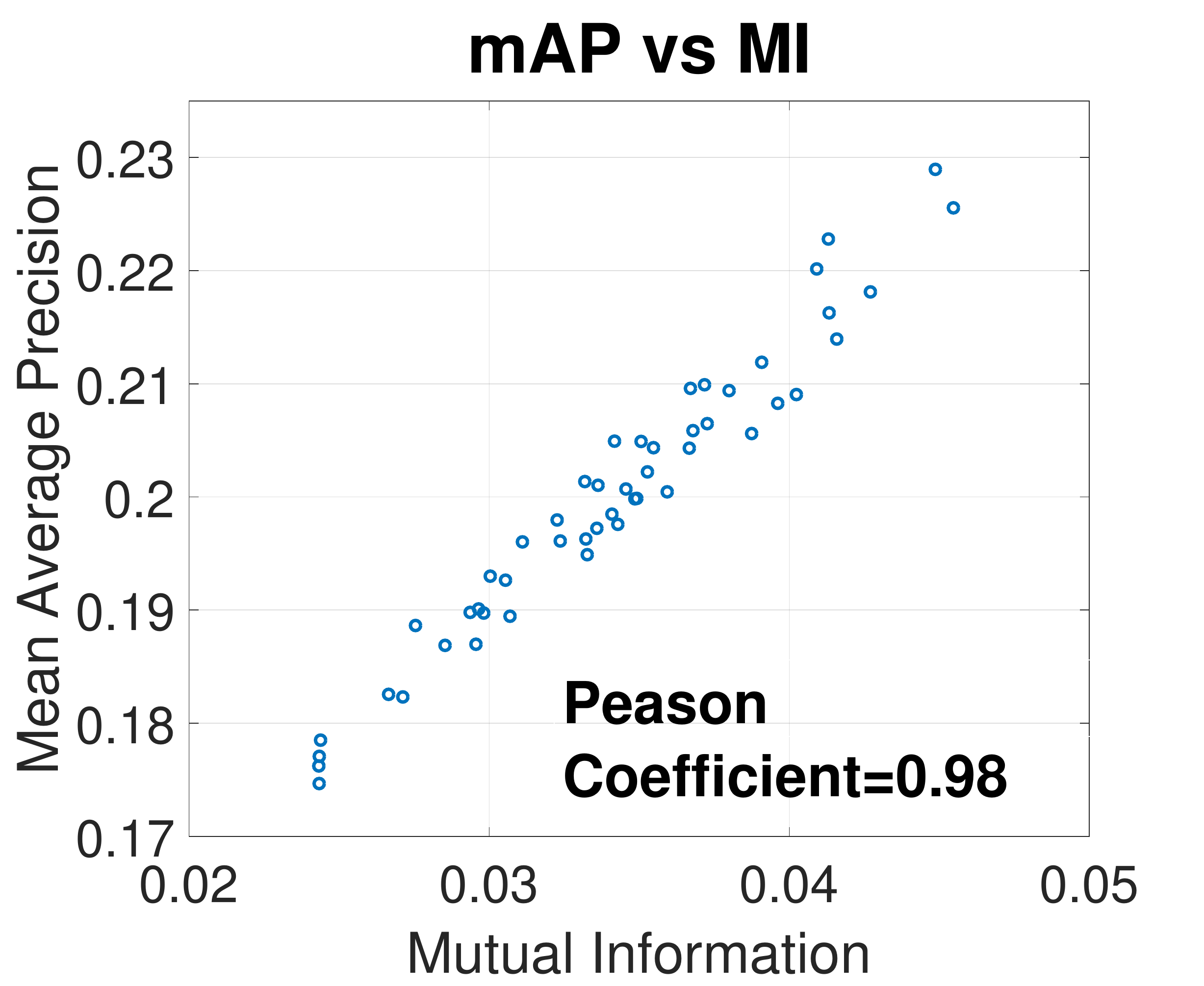}
\caption{(Left) We plot the training objective value (Equation~\ref{eq:mi_obj}) and the mAP score from the 32-bit \cifar{1} experiment. Notice that both the mutual information objective and the mAP value show similar behavior, \ie, exhibit strong correlation. (Middle) We apply min-max normalization in order to scale both measures to the same range. (Right) We conduct an additional set of experiments in which 100 instances are selected as the query set, and the rest is used to populate the hash table.
The hash mapping parameters are randomly sampled from a Gaussian, similar to LSH \cite{LSH1}. Each experiment is conducted $50$ times.
There exists strong correlation between MI and mAP as validated by the Pearson Correlation Coefficient score of 0.98.
}
\label{fig:correlation}
\end{figure*}

Retrieval results for ImageNet100 are given in Table \ref{table:imagenet}. In these experiments, we compare against DTSH, the overall best competing method in past experiments and another recently introduced deep learning based hashing method, HashNet \cite{hashnet}. 
Note that, HashNet also outperforms shallow methods such as \cite{SHK} and \cite{Shen_CVPR2015} with deep features on ImageNet100, as reported in \cite{hashnet}.
The evaluation metric is taken to be $\mathsf{mAP@1K}$ for consistency with the setup in \cite{hashnet}.
In this benchmark, \mihash~significantly outperforms both DTSH and HashNet for all embedding sizes. Notably, \mihash~achieves $6\%$ improvement over HashNet with 16-bit codes, indicating its superiority in learning high-quality compact binary codes.

To further emphasize the merits of \mihash, we consider shallow model experiments on the 22K LabelMe dataset. In this benchmark, we only consider the overall best non-deep and deep learning methods in past experiments. 
Also, to solely put emphasis on comparing the hash mapping learning objectives, all deep learning methods use a one-layer embedding on top of the \gist~descriptor. 
The \gist~descriptor is prominently used even in many recent hashing studies (\textit{e.g.}, as in \cite{FastHash} and \cite{LinIJCV2016}). 
Its usage nullifies the feature learning aspect in deep hashing techniques 
enabling a more direct comparison to non-deep hashing methods.
Still, some non-deep methods employ non-linear hash functions, such as FastHash and StructHash that use boosted decision trees. Table \ref{table:labelme} gives the results, and we can see that non-deep methods FastHash and StructHash outperform deep learning methods DPSH and DTSH on this benchmark. This indicates that the prowess of DPSH and DTSH might come primarily through feature learning. 
On the other hand, \mihash~is the best performing method across all code lengths, despite using a simpler one-layer embedding function compared to FastHash and StructHash. This further validates the effectiveness of our mutual information based objective in capturing the neighborhood structure.

\subsection{{Empirical Analysis and Ablation Studies}}

\subsubsection{Mutual Information and Ranking Metrics}

\begin{figure}
\centering
\includegraphics[trim=2.2cm 0 1.3cm 0,width=0.8\linewidth]{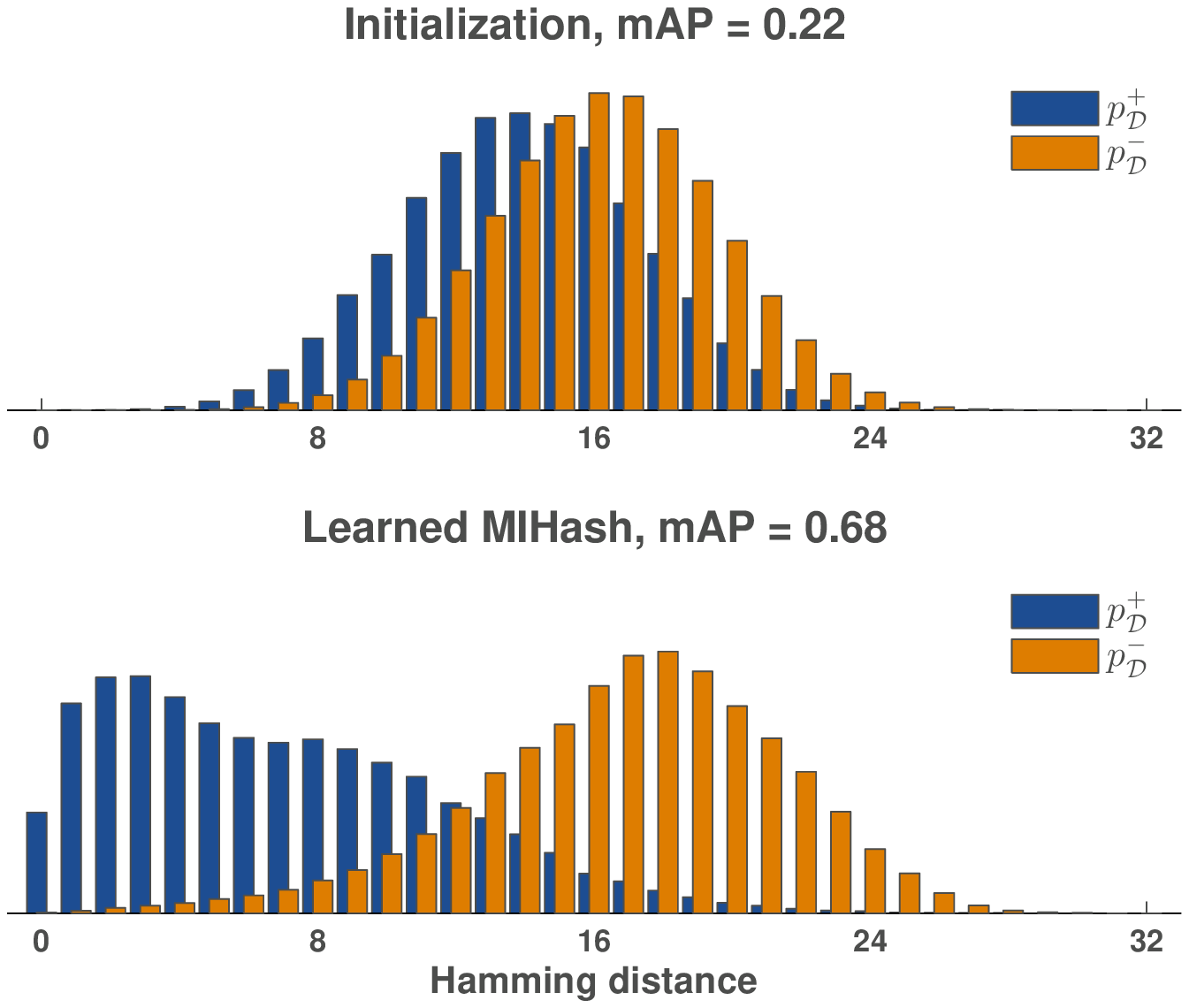}
\caption{
We plot the distributions $p_{\mathcal{D}}^+$ and $p_{\mathcal{D}}^-$, averaged on the CIFAR-10 test set, before and after learning \mihash~with a single-layer model and 20K training examples. 
Optimizing the mutual information objective substantially reduces the overlap between them, resulting in high mAP.
}
\label{fig:distance}
\end{figure}

To evaluate the performance of hashing algorithms for retrieval tasks, it is common to use ranking-based metrics, and the most notable example is mean Average Precision (mAP).
We note that there exists strong correlations between our mutual information criterion and mAP. 
Figure~\ref{fig:correlation} provides an empirical analysis on the CIFAR-10 benchmark. 
The left plot displays the training objective value as computed from Equation~\ref{eq:mi_obj} and the mAP score with respect to the epoch. 
These results are obtained from the \cifar{1} experiment with 32-bit codes, as specified in the previous section. 
Notice that both the mutual information objective and the mAP value show similar behavior, \ie, exhibit strong correlation. 
While the mAP score increases from 0.40 to 0.80, the mutual information objective increases from 0.15 to above 0.40. 
In the middle figure, we apply min-max normalization in order to scale both measures to the same range.

To further analyze the correlation between mutual information and mAP, we also conducted an additional experiment in which 100 instances are selected as the query set, and the rest are used to populate the hash table. 
The hash mapping parameters are randomly sampled from a Gaussian distribution, similar to LSH \cite{LSH1}, and each experiment is conducted 50 times. 
The right figure provides the scatter plot of mAP and the mutual information objective value. 
We can see that the relationship is almost linear, which is also validated by the Pearson Correlation Coefficient score of 0.98. 

We give an intuitive explanation to the strong correlation.
Given a query, the AP is optimized when all of its neighbors  are ranked above all of its non-neighbors in the database.
On the other hand, mutual information is optimized when the distribution of neighbor distances has no overlap with the distribution of non-neighbor distances. 
Therefore, we can see that AP and mutual information are simultaneously optimized by the same optimal solution.
Conversely, AP is suboptimal when neighbors and non-neighbors are interleaved in the ranking, so is mutual information when the distance distributions have nonzero overlap.
Although a theoretical analysis of the correlation is beyond the scope of this work, empirically we find that mutual information serves as a general-purpose surrogate metric for ranking. 

An advantage of mutual information, as we have demonstrated, is that it is suitable for direct, gradient-based optimization. 
In contrast, optimizing mAP is much more challenging as it is non-differentiable, and previous works usually resort to approximation and bounding techniques \cite{LinIJCV2016,wang2015ranking,APSVM}.

\subsubsection{Distribution Separating Effect}
To demonstrate that \mihash~indeed separates neighbor and non-neighbor distance distributions, we consider a simple experiment.
Specifically, we learn a single-layer model on top of precomputed $fc7$-layer features.
The learning is done in an online fashion, which means that each training example is processed only once.
We train such an \mihash\ model on the CIFAR-10 dataset with 20K training examples.

In Figure~\ref{fig:distance}, we plot the distributions $p_{\mathcal{D}}^+$ and $p_{\mathcal{D}}^-$, averaged on the CIFAR-10 test set, before and after learning \mihash~with 20K training examples. 
The hash mapping parameters are initialized using LSH, and lead to high overlap between the distributions, although they do not totally overlap due to the use of strong pretrained features.
After learning, the overlap is significantly reduced, with $p_{\mathcal{D}}^+$ pushed towards zero hamming distances. 
Consequently, the mAP value increases to $0.68$ from $0.22$.

\begin{figure}
\centering
\begin{subfigure}[b]{0.24\textwidth}
\centering
\includegraphics[clip, trim=2cm 6cm 2cm 6cm, width=\textwidth]{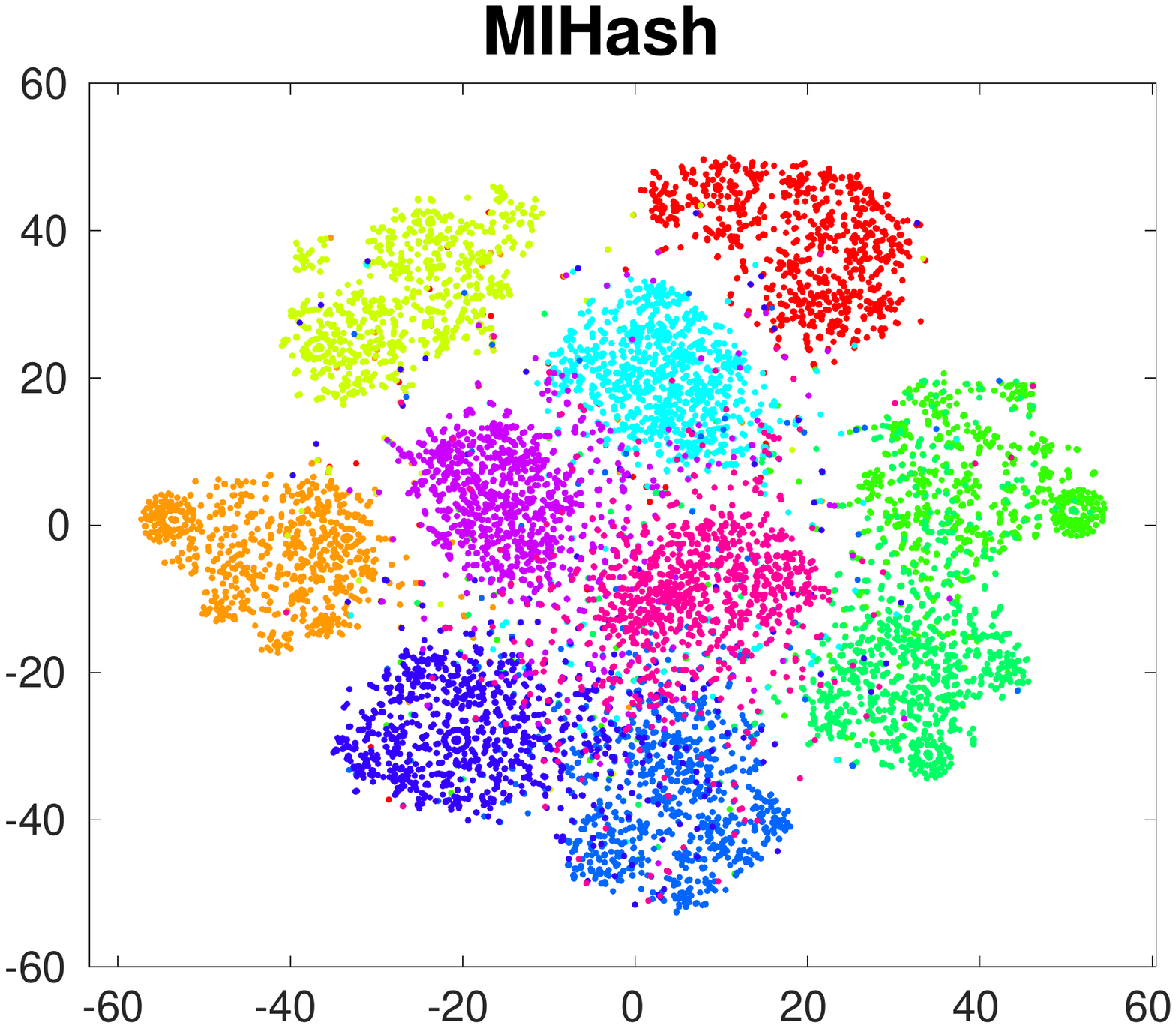}
\end{subfigure}
\begin{subfigure}[b]{0.24\textwidth}  
\centering 
\includegraphics[clip, trim=2cm 6cm 2cm 6cm, width=\textwidth]{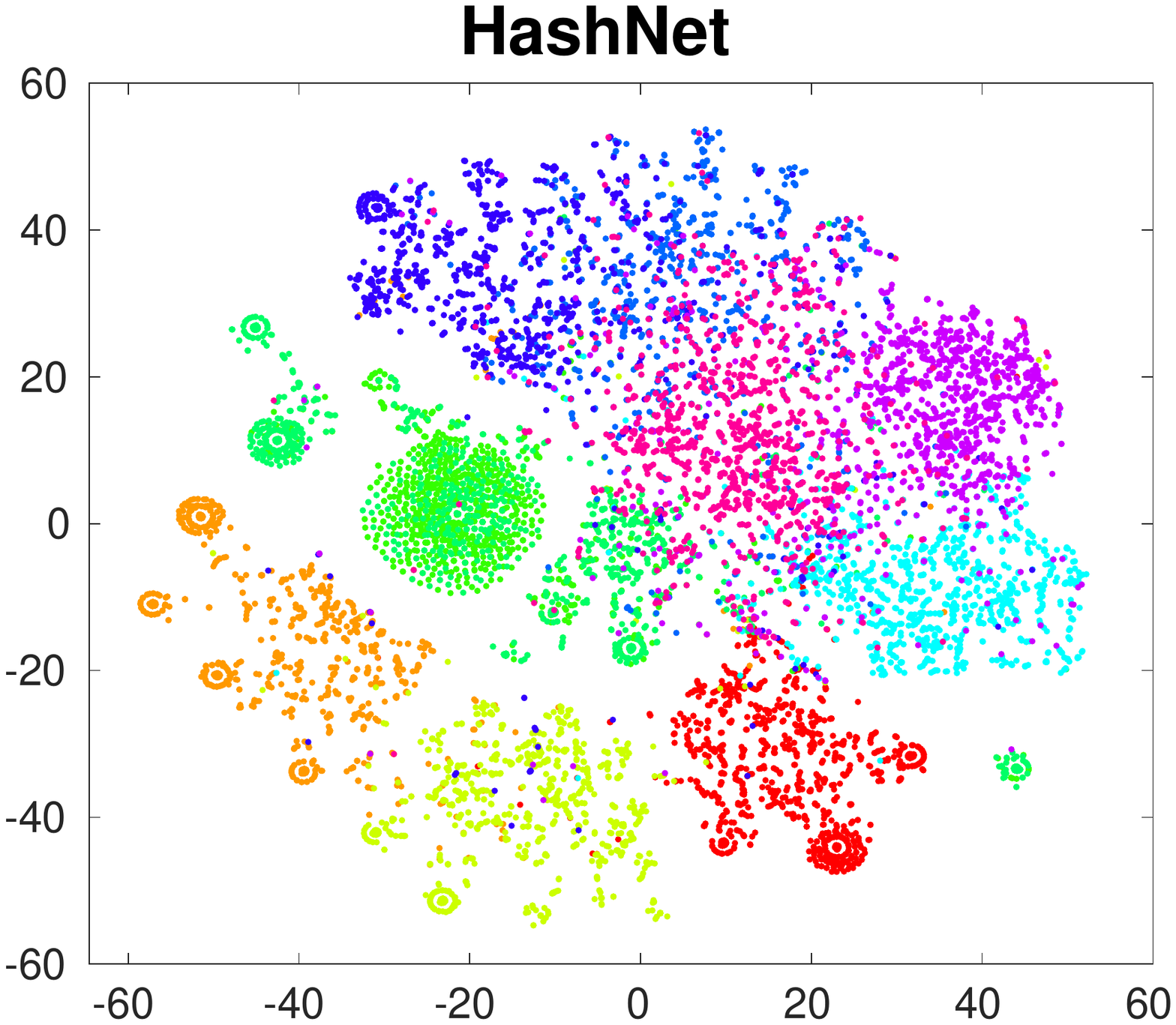}
\end{subfigure}
\vspace{-1em}
\caption
{t-SNE \cite{tsne} visualization of the $48$-bit binary codes produced by \mihash~and HashNet on ImageNet100, for a random subset of 10 different color-coded classes in the test set. \mihash~yields well-separated codes with distinct structures, opposed to HashNet, in which the binary codes have a higher overlap.} 
\label{fig:tsne}
\end{figure}

\begin{figure*}[!ht]
\centering
\includegraphics[width=.95\linewidth, trim= 0em 0em 0em 0]{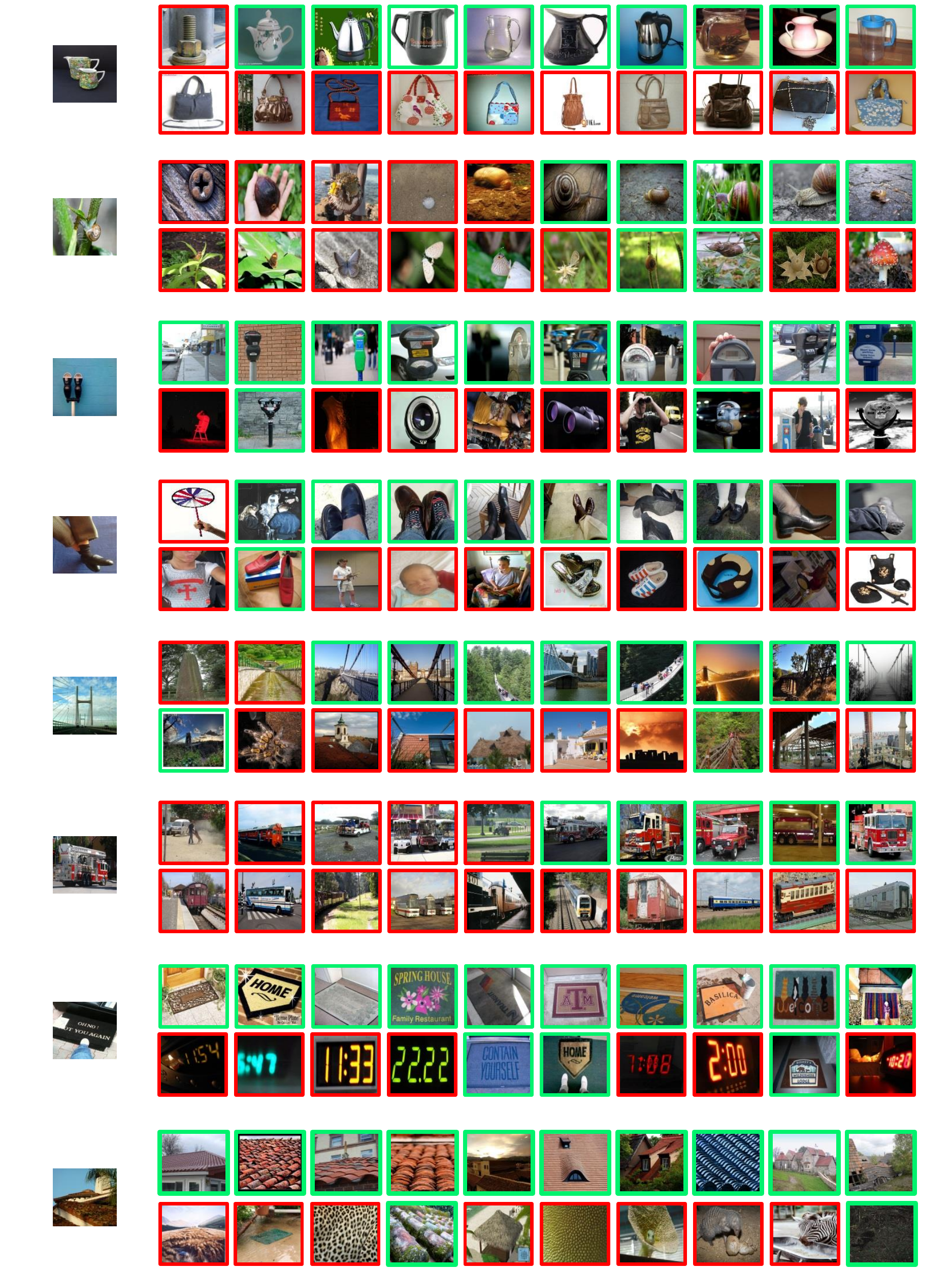}
\caption{We show  sample retrieval results from the {ImageNet100} dataset. Left: query images, right: top 10 retrieved images from \mihash~(top row) and from HashNet (bottom row). Retrieved images marked with a green border belong to the same class as the query image, while ones marked with a red border do not belong to the same class as the query image.}
\label{fig:retrieval_1}
\end{figure*} 

\subsubsection{t-SNE Visualization of the Binary Embeddings}
We also visualize the learned embeddings using t-SNE \cite{tsne}. In Figure~\ref{fig:tsne}, we plot the  visualization for 48-bit binary embeddings produced by \mihash~and the top competing method, HashNet, on ImageNet100. 
For ease of visualization, we randomly sample 10 classes from the test set.

\mihash~produces binary embeddings that separate different classes well into separate clusters. 
This is in fact predictable  from the formulation of \mihash, in which the class overlap is quantified via mutual information and minimized.
On the other hand, binary codes generated by HashNet have  higher overlap between classes.
This is also consistent with the fact that HashNet does not specifically optimize for a criterion related to class overlap, but belongs to the simpler ``affinity matching'' family of approaches.

\subsubsection{{Steepness parameter $\gamma$ and Batch Size $M$}}
We provide an ablation study on the steepness parameter $\gamma$ in Equation~19 and training minibatch size $M$. 
The experiments are conducted on the \cifar{1} benchmark with 32 bit codes. 

Generally, continuous relaxation of the binary codes introduces discrepancies between the training and testing scenarios, and is thus prone to degrading the test-time retrieval performance. 
In deep hashing studies, this issue is often mitigated by a quantization loss (\textit{e.g.}\cite{Li_IJCAI2016, Wang_ACCV2016}), or continuation methods \cite{hashnet}, or by simply keeping the binary constraints \cite{do2016ECCV}.
However, we observe the \mihash~model to be robust to the continuous relaxation: performance values are largely unaffected by variations in the $\gamma$ parameter. 
This is also true for the minibatch size parameter $M$. 
This ablation study highlights the robustness of our \mihash~formulation to the hyper-parameters.

\begin{table}
\taburulecolor{black}
\begin{tabu}{c|v|v|v|v} 
\hline
$\mathsf{VGG-F}$ &  \multicolumn{4}{c}{ $\mathsf{CIFAR-10~(mAP)}$ } \rule{0pt}{1em} \\
\hline
\multirow{1}{*}{ \emph{\textbf{Steepness}} $\gamma$ } & 1 & 5 & 10 & 20 \\
\hline
 & {0.791} & 0.768 & 0.749 & 0.776 \\
\hline
\multirow{1}{*}{ \emph{\textbf{Batch Size}} $M$ } & 64 &128 & 256 & 512 \\
\hline
& 0.771 & {0.765} & {0.791} & 0.783 \\
\hline
\end{tabu}
\caption{{Ablation study for the steepness parameter $\gamma$ and mini-batch size $M$ on the \cifar{1} benchmark with 32 bit codes.}}
\label{table:sigmoid}
\end{table}

\subsubsection{Example Retrieval Results}
In Figure~\ref{fig:retrieval_1}, we present example retrieval results for \mihash~and HashNet for several image queries from the ImageNet100 dataset. The top 10 retrievals of eight query images from eight distinct categories are presented. Correct retrievals (\ie, having the same class label as the query) are marked in green, and incorrect retrievals are in red.
In these examples, many of the retrieved images appear visually similar to the query, even if not sharing the same class label. Nevertheless, \mihash\ retrieves fewer incorrect images compared to HashNet. For example, HashNet returns bag images for the first query (image of cups), and digital-clock images for the second-to-last query (image of doormat). 

\section{Conclusion}
\label{sec:conclusion}

We take an information-theoretic approach to hashing and propose a novel hashing method, called \mihash, in this work.
It is based on minimizing neighborhood ambiguity in the learned Hamming space, which is crucial in maintaining high performance in nearest neighbor retrieval. We adopt the well-studied mutual information measure from information theory to quantify neighborhood ambiguity, and show that this measure has strong correlations with standard ranking-based retrieval performance metrics.
Then, to optimize mutual information, we take advantage of recent advances in deep learning and stochastic optimization, and parameterize our embedding functions with deep neural networks. 
We perform a continuous relaxation on the NP-hard optimization problem, and use stochastic gradient descent to optimize the networks. In particular, our formulation maximally utilizes available supervision within each minibatch, and can be efficiently implemented.
Our implementation is publicly available.

When evaluated on four standard image retrieval benchmarks, \mihash\ is shown to learn high-quality compact binary codes, and it achieves superior nearest neighbor retrieval performance compared to existing supervised hashing techniques.
We believe that the mutual information based formulation is also potentially relevant for learning real-valued embeddings, and for other applications besides image retrieval, such as few-shot learning.


%



\ifCLASSOPTIONcompsoc
  \section*{Acknowledgments}
\else
  \section*{Acknowledgment}
\fi

This research was supported in part by a BU IGNITION award, US NSF grant 1029430, and gifts from NVIDIA.

\ifCLASSOPTIONcaptionsoff
  \newpage
\fi



%



{\small
\bibliographystyle{ieee}
\bibliography{egbib}
}

%

\clearpage

\begin{IEEEbiography}[{\includegraphics[width=1in,height=1.25in,clip,keepaspectratio]{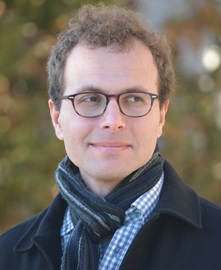}}]{Fatih Cakir}
is a Data Scientist at FirstFuel Software. He was previously a member at the Image and Video Computing Group at Boston University working with Professor Stan Sclaroff as his Ph.D. advisor. His research interests are in the fields of Computer Vision and Machine Learning.
\end{IEEEbiography}

\begin{IEEEbiography}[{\includegraphics[width=1in,height=1.25in,clip,keepaspectratio]{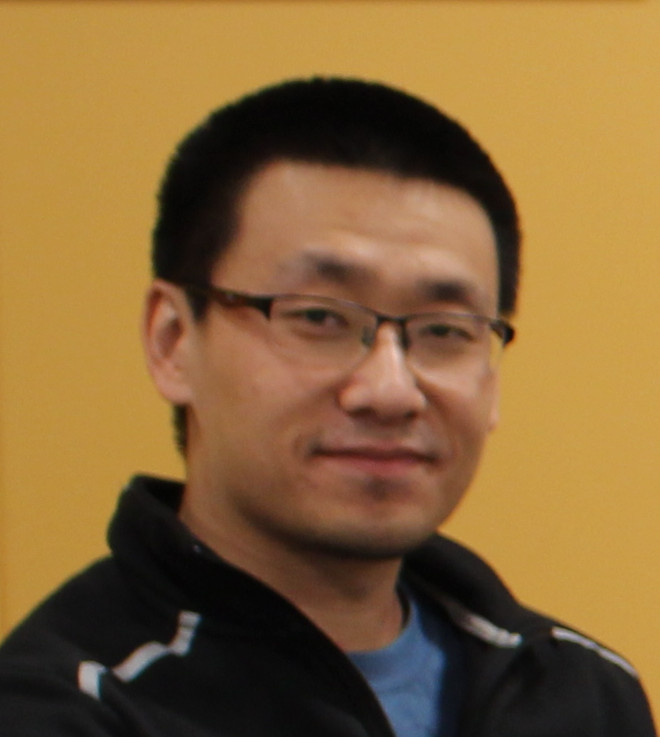}}]{\\Kun He}
is a Ph.D. candidate in Computer Science and a member of the Image and Video Computing group at Boston University, advised by Professor Stan Sclaroff. He obtained his M.Sc. degree in Computer Science from Boston University in 2013, and his B.Sc. degree (with honors) in Computer Science and Technology from Zhejiang University, Hangzhou, China, in 2010. 
He is a student member of the IEEE.
\end{IEEEbiography}


\begin{IEEEbiography}[{\includegraphics[width=1in,height=1.25in,clip,keepaspectratio]{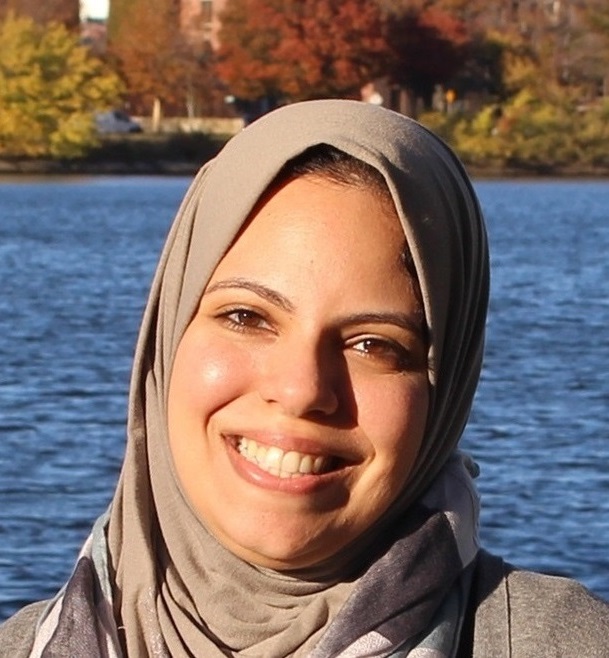}}]{\\Sarah Adel Bargal}
is a Ph.D. candidate in the Image and Video Computing group in the Boston University Department of Computer Science. She received her M.Sc. from the American University in Cairo. Her research interests are in the areas of computer vision and deep learning. She is an IBM PhD Fellow and a Hariri Graduate Fellow.
\end{IEEEbiography}

\begin{IEEEbiography}[{\includegraphics[width=1in,height=1.25in,clip,keepaspectratio]{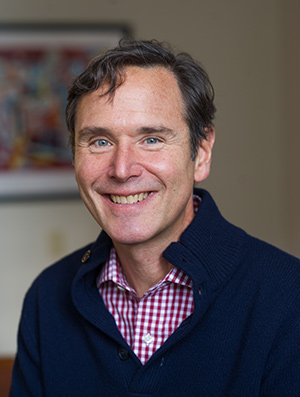}}]{Stan Sclaroff}
is a Professor in the Boston University Department of Computer Science. He received his Ph.D. from MIT in 1995. His research interests are in computer vision, pattern recognition, and machine learning. He is a Fellow of the IAPR and Fellow of the IEEE.
\end{IEEEbiography}




\end{document}